\documentclass[preprint,12pt]{elsarticle}

\usepackage{amssymb}
\usepackage{graphicx}
\usepackage{booktabs}
\usepackage{xcolor}

\usepackage{caption}
\usepackage{multirow}
\usepackage{arydshln}
\usepackage{float} % For the [H] specifier
\usepackage{algorithm} % For the algorithm environment
\usepackage{bm}
\usepackage{algorithmic} % For the algorithmic content
\usepackage{amsmath}
\usepackage{microtype}
\usepackage{dblfloatfix}
\usepackage{balance}

\definecolor{matlabgreen}{RGB}{119,172,48}
\usepackage{nomencl}

\definecolor{darkergreen}{rgb}{0.0, 0.5, 0.0}
\makenomenclature
\definecolor{darkpink}{RGB}{199,21,133}

\author[1]{Mehdi Heydari Shahna\corref{cor1}}
\author[2]{Tuomo Kivelä}
\author[1]{Jouni Mattila}

\address[1]{Faculty of Engineering and Natural Sciences, Tampere University, 33720 Tampere, Finland.}
\address[2]{Sandvik Mining and Construction Oy, Tampere, Finland.}

\cortext[cor1]{Corresponding author: Mehdi Heydari Shahna\\
E-mail address: mehdi.heydarishahna@tuni.fi }

\begin{document}

\begin{frontmatter}



\title{\textcolor{black}{Pose-Parameterized Motion Planning and CBF-QP Self-Collision Filtering for a Long-Reach Drilling Boom}}


\begin{abstract}
Long-reach drilling booms must reach successive poses without self-collision. 
\textcolor{black}{Moving from operator-supervised control toward autonomy requires collision-aware motion planning and execution. For the Sandvik SB60, this study adapts established methods by integrating pose-parameterized planning with a capsule-based control barrier function quadratic program (CBF-QP) in measured-state inverse kinematics (IK).} 
A fixed task-specific parameter set within each task generates
waypoints, detours, timed references, and chained motion without
target-specific retuning.
\textcolor{black}{The offline detour planner screens candidate waypoints using 23 selected rod-segment-to-body-region distances, whereas the online CBF-QP filters joint velocities using 14 configured capsule-pair constraints from a nine-primitive whole-body capsule model.}
\textcolor{black}{Evaluation considers two drilling tasks in a manufacturer-developed SB60 Simscape Multibody model: a five-target restricted-orientation tour and a three-target full-pose tour.}
Across several hundred thousand samples, the method produced zero IK failures, generated several detour waypoints, achieved millimetre-level mean final-position error, and recorded no sampled CBF margins below the reported thresholds.

\end{abstract}

%% Keywords

\end{frontmatter}

\section{Introduction}

\subsection{Drilling automation in underground construction}

Tunnel construction, underground mining, and rock-bolt installation depend on drilling many holes from a single equipment setup. A typical underground drilling face contains 30 to 100 drill holes, each at a specified position and angle determined by the blast pattern, ground-support design, or production requirements \citep{karlinski2008automated, nikolakopoulos2025autonomous}. Each hole's accuracy directly affects construction outcomes: hole position influences blast effectiveness in development drilling, hole angle affects bolt installation reliability in ground support, and accumulated positioning errors can propagate to over-break, under-break, and rework.

The equipment performing these operations is a hydraulic boom manipulator. Modern drilling rigs may carry one or more booms, each comprising
multiple actuated joints that position a long drilling rod, typically
several meters in length \citep{sandvikDT1132i}. 

These booms are kinematically redundant in general, with more actuated coordinates than the six required for a general tip pose, although procedure-scheduled joints may not be available for redundancy resolution in every operating mode. Their long reach and coupled geometry create a difficult control problem: transitions can bring the rod, boom housing, and carrier into close proximity, direct paths may require detours, and joint capabilities are anisotropic across the kinematic chain.

Despite advances in automated boom positioning and pre-planned drilling-pattern execution, many boom transitions remain operator-supervised. In such workflows, the operator monitors boom motion and intervenes when interference is observed \citep{karlinski2008automated,sandvikDT1132i}.

Increasing the autonomy of these operations is a recognized goal in
construction-automation research because it can reduce cycle time and
mitigate operators' exposure to hazards in confined drilling
environments
\citep{nikolakopoulos2025autonomous,kulz2025holistic,
kim2025sharedAutonomy,kostavelis2024robetarme}.

\subsection{Current production practice and its limits}

Three common limitations in operator-supervised drilling workflows motivate further automation.
\begin{itemize}
    \item \textit{Pose-dependent motion preparation.} Workflows based on manually defined trajectories or pose-specific waypoints require updates when the drilling face or rig setup changes \citep{karlinski2008automated,sandvikDT1132i}.
    
    \item \textit{Operator-monitored collision avoidance.} In workflows without automated collision checking or safety filtering, the operator monitors proximity between the rod, boom, and carrier and intervenes when necessary.
    
    \item \textit{Conservative motion-rate scaling.} Reduced motion rates provide more time for operator intervention but
increase transition duration.
\end{itemize}

These limitations motivate a workflow that links drilling-pose specifications to collision-aware motion and feedback safety filtering. \textcolor{black}{The following review positions the present work within established
motion-planning, geometric collision-checking, and CBF-based
safety-control research. This study adapts and integrates these
components for closed-loop execution on an SB60-class boom.}

\subsection{Motion generation and safety filtering: related work and positioning}

Manipulator motion planning, collision avoidance, and execution-time
safety filtering are established but distinct research threads \cite{HeydariShahna2024DeepRL}.
Sampling-based planners and trajectory optimizers search for
collision-free nominal paths, for example through rapidly exploring
random tree (RRT) variants or stochastic trajectory optimization for
motion planning (STOMP) \citep{kalakrishnan2011stomp}.
Ragaglia et al.~\citep{ragaglia2017cartesian} combined Cartesian-space
RRT planning, inverse kinematics (IK), capsule collision checking,
path smoothing, and kinematic time scaling for autonomous
construction-machine motion.

\textcolor{black}{Geometric path construction and temporal parameterization are
distinct. Alizadeh Kolagar et al.~\citep{AlizadehKolagar2025JBTG}
combined deep reinforcement learning (DRL) motion commands with a
jerk-bounded trajectory generator and precomputed safe-velocity zones
to respect joint-level kinematic constraints.
Section~\ref{sec:reference_trajectory_generation} specifies the
seventh-order timing law and controller-grid discretization used here
and clarifies that the nominal duration estimate does not certify the
nonlinear IK solution or the filtered plant trajectory
\citep{biagiotti2008trajectory,berscheid2021ruckig}.} Artificial potential fields (APFs) provide reactive local corrections \citep{khatib1985realTime}. Control barrier function quadratic programs (CBF-QPs) instead modify a nominal control input subject to model-based inequality constraints and provide forward-invariance results only under their stated assumptions \citep{rauscher2016constrained,ames2017cbfqp,ames2019cbfTheory}. \textcolor{black}{These standard forward-invariance results concern a continuous-time model. Enforcing a continuous-time CBF condition only at controller samples does not, in general, establish intersample invariance; sampled-data guarantees require additional intersample bounds or sampled-data barrier conditions \citep{breeden2022sampledCBF}. Accordingly, this study states a conditional continuous-time result for the velocity-level capsule model and treats the simulation with a \(5~\mathrm{ms}\) controller period and
second-order command filtering as empirical clearance evidence, not as an unconditional sampled-data or physical-machine guarantee.} An offline planner shapes a nominal route according to its own
geometric model, whereas the runtime filter acts according to its
separate configured constraint model.

Work on heavy hydraulic and drilling manipulators already covers several parts of the present pipeline. Kivelä et al. \citep{kivela2017generic} optimized the kinematic structure and placement of a heavy underground drilling manipulator at design time.
Their subsequent collision-avoidance controller \citep{kivela2018redundant} used point-cloud-based extended oriented bounding boxes (EOBBs),
damped least-squares (DLS) redundancy resolution, null-space projection, and an APF for real-time obstacle and self-collision avoidance on a heavy hydraulic tunneling manipulator. Complementary actuator-level work on heavy-duty manipulators driven by
electromechanical linear actuators (EMLAs) has used observer-based modular robust control to track
planned joint trajectories under modeling uncertainty and disturbances
\cite{HeydariShahna2025Observer}. More recent drilling-specific studies have addressed hole-sequence and arm-motion planning directly. Li et al. \citep{li2023drillingPath} combined vehicle-mounted scanning, local collision-aware arm motion, and traveling-salesman sequencing of drilling positions. Zhang et al. \citep{zhang2023improvedRRT} combined an improved RRT, an APF, greedy pruning, and B-spline smoothing for an anchor-drilling arm. Sun et al. \citep{sun2024drillAnchor} used an improved bidirectional RRT (Bi-RRT) and APF together with cylinder- and sphere-based collision checks for a drill-and-anchor material-handling arm.

Nie et al.~\citep{nie2025rockDrilling} integrated an oriented-vector
variant of the bidirectional RRT algorithm (v-BI-RRT) with oriented
bounding-box (OBB) and capsule envelopes for self-collision-free route planning on a redundant rock-drilling manipulator. Xu and Lin \citep{xu2025mssRRT} combined multi-strategy RRT*, learned sampling, cylindrical collision envelopes, and B-spline smoothing for a large multi-arm rock-drilling robot. Cui et al. \citep{cui2025multiarm} combined genetic path optimization, an improved APF, and B-spline trajectory generation for cooperative tunnel-drilling arms. \textcolor{black}{The evaluation scales reported by Li et al.~\citep{li2023drillingPath}
and Cui et al.~\citep{cui2025multiarm} also place the two complementary
validation tasks in context: Li et al. report 57--62 reachable drilling points per cycle across eight cycles, and Cui et al. allocate 40, 52, and 40 holes to three arms. The tasks and platforms are different, so these counts are not a performance benchmark, but they show why the multi-target tour below must be described as an initial feasibility case rather than production-pattern validation.}

CBF-based manipulator safety is likewise established. Rauscher et al. \citep{rauscher2016constrained} formulated constrained robot control as a CBF-QP. Khazoom et al. \citep{khazoom2022humanoid} used sphere and capsule signed distances in a CBF whole-body controller for humanoid self-collision avoidance. Morton and Pavone \citep{morton2025oscbf} enforced whole-body, self-collision, workspace, and singularity constraints with operational-space CBFs at real-time rates. Shi et al. \citep{shi2025spaceCBF} combined capsule-shaped safety envelopes, high-order CBF filtering, and a replanning mechanism for a free-flying space robot. In construction robotics, Kim et al. \citep{kim2025sharedAutonomy} reported a shared-autonomy overhead-work system using a neural configuration-space barrier for drilling, anchoring, and bolting. The system of Kim et al.~\citep{kim2025sharedAutonomy} is an initial
prototype on a lighter platform and differs from autonomous underground drilling, but it is directly relevant to barrier-based construction-manipulator safety.

\begin{table*}[h!]
\centering
\caption{\textcolor{black}{Positioning relative to relevant studies.}}
\label{tab:literature_positioning}
\scriptsize
\begin{tabular}{@{}p{2.1cm}p{4.7cm}p{4.7cm}@{}}
\toprule
Study group & Reported method and application & Relation to the present study \\
\midrule
Kivelä et al. \citep{kivela2017generic,kivela2018redundant} & Heavy underground/hydraulic manipulator design optimization and real-time DLS/null-space/APF collision avoidance using point-cloud EOBBs. & Closest prior work on the same manipulator class; it uses a different local safety mechanism and reports a single collision-avoidance maneuver. \\
\midrule
Ragaglia et al.; Li et al.; Zhang et al.; Sun et al. \citep{ragaglia2017cartesian,li2023drillingPath,zhang2023improvedRRT,sun2024drillAnchor} & Construction or drilling motion generation using Cartesian RRT, RRT/Bi-RRT, APFs, geometric collision checks, path smoothing, and drilling-position sequencing. & Establish construction- and drilling-specific planning, including capsule collision checks; they do not use the present SB60 velocity-level capsule-pair CBF-QP. \\
\midrule
Nie et al.; Xu and Lin; Cui et al. \citep{nie2025rockDrilling,xu2025mssRRT,cui2025multiarm} & Capsule/OBB or cylindrical collision models combined with sampling, learning, genetic optimization, APFs, and smoothing for single- or multi-arm rock-drilling manipulators. & Closest planning comparators; they establish that capsule-based collision-aware motion generation
for drilling arms is not itself novel. \\
\midrule
Rauscher et al.; Khazoom et al.; Morton and Pavone \citep{rauscher2016constrained,khazoom2022humanoid,morton2025oscbf} & CBF-QP or whole-body CBF control for constrained manipulation, including capsule-based self-collision constraints and large sets of operational-space constraints. & Establish the general CBF-QP and whole-body self-collision methodology on non-drilling platforms. \\
\midrule
Shi et al.; Kim et al. \citep{shi2025spaceCBF,kim2025sharedAutonomy} & Capsule-HOCBF tracking with replanning for a space robot, and neural-barrier planning/control for shared-autonomy overhead construction work. & Show that barrier filtering has already been combined with replanning or construction tasks, but with different platforms, autonomy levels, dynamics, and validation goals. \\
\midrule
Present study & Pose-parameterized SB60 waypoint/detour generation using 23 selected rod-segment-to-body-region distances, chained execution, and a measured-joint-state velocity-level CBF-QP imposing 14 configured whole-body capsule-pair constraints. & Application-specific integration evaluated in two tasks using a manufacturer-developed model: a five-target restricted-orientation tour for planner--CBF interaction characterization and a three-target full six-degree-of-freedom (6-DOF) tour for continuous pose-tracking
assessment.\\
\bottomrule
\end{tabular}
\end{table*}

\textcolor{black}{The literature therefore shows that the individual components of the framework, including multi-pose planning, capsule collision models, CBF-QP filtering, and model-based validation, have clear precedents.} The contribution of this study is the SB60-specific engineering integration of a pose-parameterized motion layer with a measured-state capsule-pair CBF-QP, together with a transparent model-based assessment of how the planner and filter interact. \textcolor{black}{Relative to drilling-specific RRT/APF methods, the distinguishing feature is retention of an execution-time CBF constraint layer. Relative to general manipulator CBF work, the distinguishing feature is adaptation to the geometry, joint limits, long-rod collision pairs, and chained drilling-pose interface of a heavy hydraulic boom.}

\textcolor{black}{The comparison in Table~\ref{tab:literature_positioning} is functional rather than a common-benchmark performance comparison. Consequently, the ablations in
Section~\ref{sec:experimental_setup} compare only components of the
proposed implementation.}

Capsules, cylinders with hemispherical end caps, are a standard abstraction for whole-body manipulator collision modeling \citep{safeea2019capsules}. In this work, the capsule-pair distance is computed in closed form, while the distance gradient used in the CBF rows is evaluated by finite differences with respect to the joint variables. \textcolor{black}{This provides the distance and local linearization needed for the CBF-QP constraints; neither capsule geometry nor CBF theory is claimed as a standalone methodological novelty.}

\subsection{Contributions and validation scope}

Building on the SB60 kinematic foundations of Kivelä et al. \citep{kivela2017generic} and the collision-avoidance objectives \citep{kivela2018redundant}, the contributions are:

\begin{enumerate}
 
    \item An SB60-specific, pose-parameterized motion layer that accepts a 6-DOF drilling-pose specification and combines adaptive \(z\)-lift, waypoint chaining, geometric detours, seventh-order time scaling, and inter-hop state propagation. \textcolor{black}{The normalized timing profile, analytical peak-derivative factors, task-dependent duration rules, task-specific timing scales, joint command bounds, and upward rounding to the controller grid with sampling period
\(T_s=5~\mathrm{ms}\) are reported for reproducibility. Within each task, the same settings are reused across all targets and ablation modes without target-specific retuning.}

    \item A measured-joint-state velocity-level CBF-QP implementation for the SB60, with CBF inequalities constructed from 14 configured capsule-pair distances in a nine-primitive whole-body capsule model. A separate reduced planning model evaluates 23 selected rod-segment-to-body-region distances for offline waypoint-risk assessment and detour scoring. The larger number of planner distances results from subdividing the rod and retaining multiple local segment-to-region checks; it does not indicate broader safety coverage or define the online CBF safe set. \textcolor{black}{The platform-specific contribution is the geometric mapping and integration into the IK loop; the model-conditional scope is stated in Section~\ref{sec:cbf_qp_ik}.}

    \item Internal component ablations characterize planner--CBF interaction in Task~1, while Task 2 evaluates continuous full 6-DOF regulation under nominally collision-clear transit geometry.

\item \textcolor{black}{An initial software-in-the-loop study comprising two complementary tasks in a manufacturer-developed SB60 Simscape Multibody model: a five-target restricted-orientation tour evaluated over \(299{,}757\) endpoint-inclusive controller samples (\(1498.780~\mathrm{s}\)) and a three-target full 6-DOF tour evaluated over \(73{,}731\) samples. The evaluation covers tracking, solver failures, detour generation, joint-limit use, and the sampled 14-pair CBF margin.}

\end{enumerate}

\textcolor{black}{Here, ``multi-pose'' denotes chained execution of distinct target poses. The results provide application-specific feasibility evidence for the evaluated model and pose set.}

The remainder of the paper is organized as follows. Section~2 describes the SB60 boom and the two drilling-pose tasks. Section~3 presents the motion-generation method, including the task-dependent joint scheduling, CBF-QP formulation, detour planner, and tour-mode chaining. Section~4 reports the model-based validation and ablation results. Section~5 concludes the paper.

\section{System description}
\label{sec:plant_model}

This section describes the SB60 manipulator, its kinematic model and joint capabilities, and the drilling-pose specifications used in the two evaluations.

\begin{figure}[h!]
\hspace*{-0.0cm} % Adjust the value as needed
\centering
\scalebox{0.75}{\includegraphics[trim={0cm 0.0cm 0.0cm 0cm},clip,width=\columnwidth]{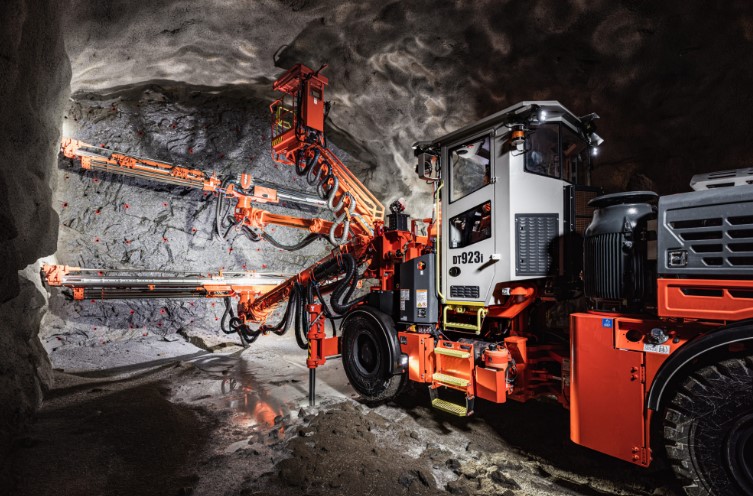}}
\caption{Sandvik jumbo with dual drilling booms \citep{sandvik_underground_drill_rigs}.}
\label{fig:dt923i_jumbo}
\end{figure}

The Sandvik SB60 is a hydraulic universal roll-over boom used in mechanized tunneling, drifting, cross-cutting, rock-bolt installation, and production drilling in cut-and-fill and room-and-pillar mining \citep{kivela2017generic}. Mounted on a carrier, it positions a drilling rod at the rock face. Figure~\ref{fig:dt923i_jumbo} shows the dual-boom DT923i as an example of the carrier-and-boom arrangement; the pictured machine uses a different boom model and is not the evaluated SB60 system \citep{sandvik_underground_drill_rigs}.

\begin{figure}[h!]
\hspace*{-0.0cm} % Adjust the value as needed
\centering
\scalebox{0.75}{\includegraphics[trim={0cm 0.0cm 0.0cm 0cm},clip,width=\columnwidth]{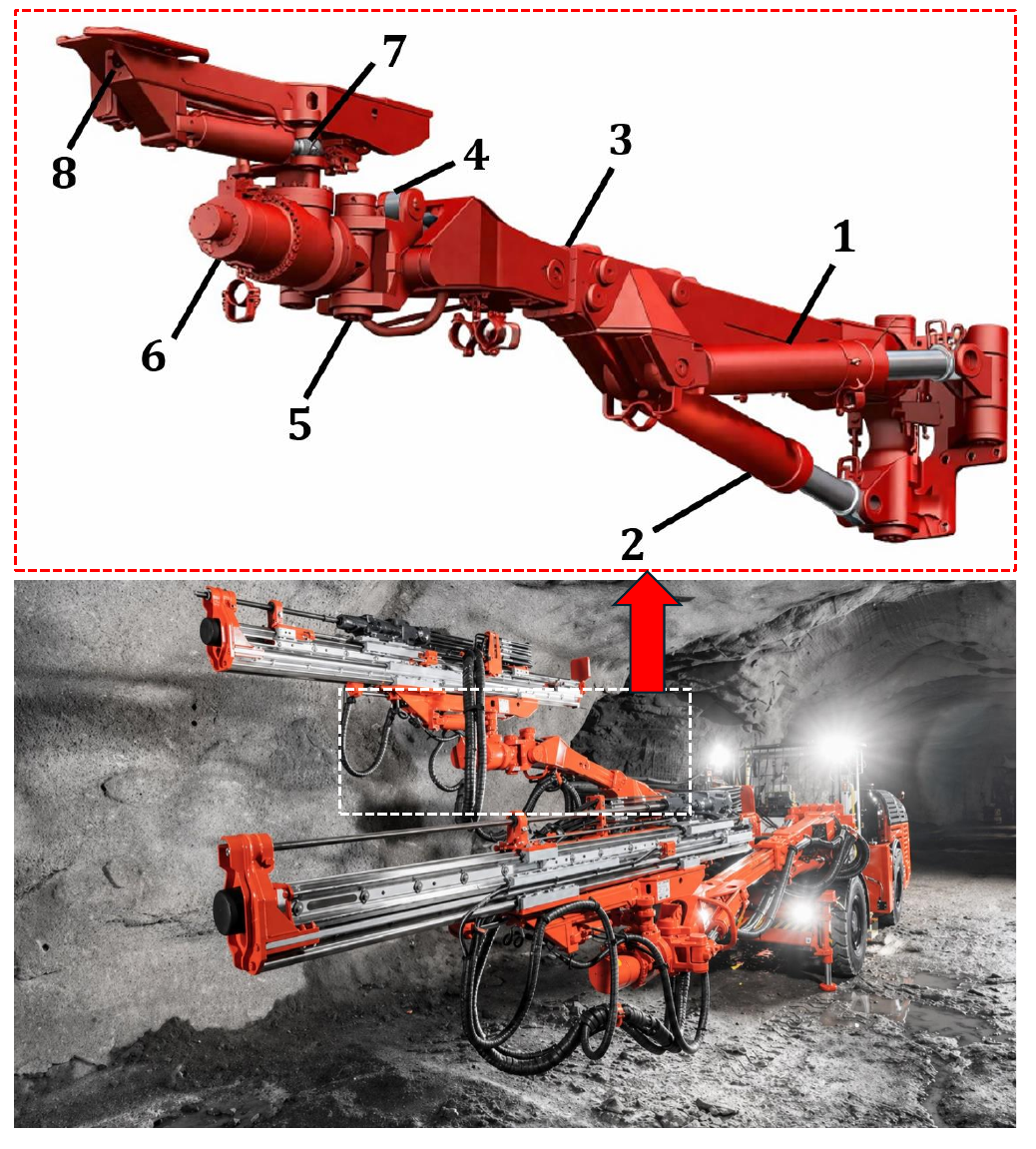}}
\caption{Sandvik SB60 boom mounted on a tunneling-jumbo carrier.
Labels 1--8 correspond to \(q_1\)--\(q_8\) in the order defined in
the text \citep{kivela2017generic}. The lower image locates the boom on the carrier, and the
arrow links the highlighted boom to the enlarged view.}
\label{fig:sb60_mobility}
\end{figure}

As shown in Fig.~\ref{fig:sb60_mobility}, the SB60 has eight degrees
of mobility (DOM): six revolute joints and two prismatic joints.
Following the joint numbering of Kivelä et al.~\citep{kivela2017generic},
the joints are as follows: \(q_1\), boom swing; \(q_2\), boom lift;
\(q_3\), boom zoom (prismatic); \(q_4\), feed tilt; \(q_5\), feed
swing; \(q_6\), feed roll; \(q_7\), feed tilt for bolting; and
\(q_8\), feed extension (prismatic). The drilling rod is mounted at
the feed end, after \(q_8\), and points along the \(+X\) axis of the
end-effector body frame in the adopted convention.

\textcolor{black}{The availability of the feed-roll and bolting-tilt joints depends on the evaluated task. The task-dependent joint partitions are
\[
\begin{aligned}
\mathcal{J}_{\mathrm{free}}^{(1)}
&=\{1,2,3,4,5,8\},
&
\mathcal{J}_{\mathrm{exact}}^{(1)}
&=\{6,7\},\\
\mathcal{J}_{\mathrm{free}}^{(2)}
&=\{1,2,3,4,5,6,8\},
&
\mathcal{J}_{\mathrm{exact}}^{(2)}
&=\{7\}.
\end{aligned}
\]
In Task~1, \(q_6\) and \(q_7\) follow procedure-scheduled references and are unavailable for Cartesian redundancy resolution. In Task~2, \(q_7\) follows an exact scheduled reference, whereas \(q_6\) remains a freely optimized coordinate with a soft posture reference, allowing the QP to regulate the complete Cartesian orientation. At the drilling targets, the scheduled values are \(q_6=0^\circ\) and \(q_7=90^\circ\) in Task~1, whereas \(q_7=90^\circ\) in Task~2.}

With the rod mounted, the manipulator tip reaches approximately \(7~\mathrm{m}\) from the carrier, and the evaluated task requires millimeter-scale positioning at multiple drilling poses across the tunnel face \citep{kivela2017generic,sandvikDT1132i}.

The mechanism's kinematic flexibility also creates collision-avoidance challenges: a direct transition between two reachable drilling poses can bring the rod or boom housing close to other parts of the boom or carrier. Detour waypoints may therefore be required to avoid self-collision \citep{kivela2018redundant}.

The boom kinematic model uses the modified Denavit--Hartenberg (DH)
parameterization and fixed link values reported by Kivelä
et al.~\citep[Table~1]{kivela2017generic}.
Table~\ref{tab:dh_parameters} gives the symbolic joint structure used
in this work.

\textcolor{black}{We define the generalized joint vector as
\[
\mathbf{q}
=
[q_1,q_2,q_3,q_4,q_5,q_6,q_7,q_8]^\top
=
[\theta_1,\theta_2,d_{3j},\theta_4,\theta_5,
 \theta_6,\theta_7,d_8]^\top .
\]
The constant angular offsets shown in Table~\ref{tab:dh_parameters}
belong to the DH transformation and are not part of the actuated
coordinates \(q_i\).}

The fixed geometric parameters are
\(a_1\), \(a_2\), \(a_4\), \(d_{3p}\), \(d_4\), \(d_6\), and \(d_7\).
The variables \(d_{3j}\) and \(d_8\) are prismatic coordinates, while
the remaining actuated coordinates are revolute.

\begin{table}[h]
\centering
\caption{Symbolic modified-DH structure of the SB60.
\textcolor{black}{Length parameters are expressed in meters and
angular parameters in radians. Joint variables are shown in bold.}}
\label{tab:dh_parameters}
\begin{tabular}{ccccc}
\toprule
Frame \(i\) & \(a_{i-1}\) & \(d_i\) & \(\alpha_{i-1}\) & \(\theta_i\) \\
\midrule
1 & \(0\)   & \(0\)                              & \(0\)      & \(\boldsymbol{\theta_1}\) \\
2 & \(a_1\) & \(0\)                              & \(\pi/2\)  & \(\boldsymbol{\theta_2}+\pi/2\) \\
3 & \(a_2\) & \(d_{3p}+\boldsymbol{d_{3j}}\)     & \(\pi/2\)  & \(0\) \\
4 & \(0\)   & \(d_4\)                            & \(\pi/2\)  & \(\boldsymbol{\theta_4}+\pi/2\) \\
5 & \(a_4\) & \(0\)                              & \(\pi/2\)  & \(\boldsymbol{\theta_5}+\pi/2\) \\
6 & \(0\) & \(d_6\) & \(\pi/2\) & \(\boldsymbol{\theta_6}\) \\
7 & \(0\) & \(d_7\) & \(-\pi/2\) & \(\boldsymbol{\theta_7}\) \\
8 & \(0\)   & \(\boldsymbol{d_8}\)               & \(\pi/2\)  & \(0\) \\
\bottomrule
\end{tabular}

\par\smallskip
\begin{minipage}{0.96\columnwidth}
\footnotesize
\textcolor{black}{\textit{Note:} The optimization retains all eight
joint-velocity entries. In Task~1, \(q_6\) and \(q_7\) are imposed as
exact procedure-scheduled coordinates, leaving six freely optimized
velocities. In Task~2, only \(q_7\) is exact-scheduled; \(q_6\) remains
freely optimized with a soft posture reference, leaving seven freely
optimized velocities. All eight coordinates remain in the kinematic
and collision calculations.}
\end{minipage}
\end{table}

The representative joint ranges and capability limits used in this work are summarized in Table~\ref{tab:joint_capabilities}. Due to confidentiality constraints, these values should not be interpreted as exact specifications of the physical system. Because the chain contains both revolute and prismatic coordinates,
their raw limits are not directly comparable across joint types. In
Task~1, the endpoint-IK duration rule evaluates each freely optimized
joint displacement against that joint's own velocity and acceleration
limits. In Task~2, segment duration is instead calculated from
Cartesian position-component and shortest-arc orientation
displacements using the limits reported in
Table~\ref{tab:controller_planner_parameters}. The joint limits in this
table continue to bound the QP commands in both tasks. Here,
\(v_{\max}\) and \(a_{\max}\) denote the maximum commanded joint
velocity and acceleration, respectively.

\begin{table*}[t]
\centering
\caption{Representative joint ranges and task-dependent command bounds used in the evaluation.}
\label{tab:joint_capabilities}
\small
\begin{tabular}{llrrrr}
\toprule
& & \multicolumn{2}{c}{Task~1} & \multicolumn{2}{c}{Task~2} \\
Joint & Range & \(v_{\max}\) & \(a_{\max}\) & \(v_{\max}\) & \(a_{\max}\) \\
\midrule
\(q_1\) & \(\pm45^\circ\) &
\(10^\circ/\mathrm{s}\) & \(40^\circ/\mathrm{s^2}\) &
\(10^\circ/\mathrm{s}\) & \(40^\circ/\mathrm{s^2}\) \\

\(q_2\) & \([-25^\circ,55^\circ]\) &
\(8^\circ/\mathrm{s}\) & \(35^\circ/\mathrm{s^2}\) &
\(8^\circ/\mathrm{s}\) & \(35^\circ/\mathrm{s^2}\) \\

\(q_3\) & \([0,1.2]~\mathrm{m}\) &
\(0.06~\mathrm{m/s}\) & \(0.25~\mathrm{m/s^2}\) &
\(0.06~\mathrm{m/s}\) & \(0.25~\mathrm{m/s^2}\) \\

\(q_4\) & \([-55^\circ,35^\circ]\) &
\(12^\circ/\mathrm{s}\) & \(60^\circ/\mathrm{s^2}\) &
\(8^\circ/\mathrm{s}\) & \(35^\circ/\mathrm{s^2}\) \\

\(q_5\) & \(\pm45^\circ\) &
\(12^\circ/\mathrm{s}\) & \(60^\circ/\mathrm{s^2}\) &
\(10^\circ/\mathrm{s}\) & \(40^\circ/\mathrm{s^2}\) \\

\(q_6\) & \(\pm179^\circ\) &
scheduled & scheduled &
\(20^\circ/\mathrm{s}\) & \(80^\circ/\mathrm{s^2}\) \\

\(q_7\) & \([0^\circ,90^\circ]\) &
scheduled & scheduled &
\(8^\circ/\mathrm{s}\) & \(35^\circ/\mathrm{s^2}\) \\

\(q_8\) & \([0,1.0]~\mathrm{m}\) &
\(0.10~\mathrm{m/s}\) & \(0.40~\mathrm{m/s^2}\) &
\(0.06~\mathrm{m/s}\) & \(0.25~\mathrm{m/s^2}\) \\
\bottomrule
\end{tabular}
\end{table*}

\section{Method}
\label{sec:method}

This section presents the target-agnostic reference generator, capsule collision models, CBF-QP IK solver, geometric detour planner, and tour-mode driver.

\subsection{Reference-trajectory generation}
\label{sec:reference_trajectory_generation}

In the implementation, the orientation waypoints lie on a single
shortest-path quaternion geodesic from \(Q_{\mathrm{start}}\) to
\(Q_{\mathrm{goal}}\). Their interpolation parameters are determined by
the normalized cumulative Cartesian arc length:
\begin{equation}
s_k
=
\frac{
\displaystyle\sum_{m=1}^{k-1}
\left\|
\mathbf{p}_{m+1}-\mathbf{p}_{m}
\right\|_2
}{
\displaystyle\sum_{m=1}^{3}
\left\|
\mathbf{p}_{m+1}-\mathbf{p}_{m}
\right\|_2
},
\qquad
\mathbf{p}_k\in
\left\{
\mathbf{p}_{\mathrm{start}},
\mathbf{p}_{\mathrm{escape}},
\mathbf{p}_{\mathrm{mid}},
\mathbf{p}_{\mathrm{goal}}
\right\}.
\label{eq:orientation_waypoint_progress}
\end{equation}
Accordingly,
\begin{equation}
\begin{aligned}
Q_{\mathrm{escape}}
&=
\operatorname{SLERP}
\left(
Q_{\mathrm{start}},
Q_{\mathrm{goal}};
s_{\mathrm{escape}}
\right),\\
Q_{\mathrm{mid}}
&=
\operatorname{SLERP}
\left(
Q_{\mathrm{start}},
Q_{\mathrm{goal}};
s_{\mathrm{mid}}
\right).
\end{aligned}
\label{eq:intermediate_orientation_waypoints}
\end{equation}
The sign of \(Q_{\mathrm{goal}}\) is reversed when necessary to ensure
shortest-path interpolation. Consecutive orientation waypoints are then
interpolated by spherical linear interpolation (SLERP)
\citep{chen2016real,dong2020research}.
Let
\[
d_h=
\sqrt{
(x_{\mathrm{goal}}-x_{\mathrm{start}})^2+
(y_{\mathrm{goal}}-y_{\mathrm{start}})^2
}.
\]
The implemented additional lift and safe transit height are
\begin{equation}
\Delta z_{\mathrm{lift}}
=
\min\!\left\{
0.45,
\max\!\left[
0.20,\,
0.20+0.03d_h
\right]
\right\},
\qquad
z_{\mathrm{safe}}
=
\max(z_{\mathrm{start}},z_{\mathrm{goal}})
+
\Delta z_{\mathrm{lift}}.
\label{eq:adaptive_z_lift}
\end{equation}
The escape and midpoint positions are
\[
\mathbf p_{\mathrm{escape}}
=
[x_{\mathrm{start}},y_{\mathrm{start}},z_{\mathrm{safe}}]^\top,
\qquad
\mathbf p_{\mathrm{mid}}
=
\left[
\frac{x_{\mathrm{start}}+x_{\mathrm{goal}}}{2},
\frac{y_{\mathrm{start}}+y_{\mathrm{goal}}}{2},
z_{\mathrm{safe}}
\right]^\top.
\]
Thus, the additional lift is limited to
\(0.20\)--\(0.45~\mathrm{m}\).

For a segment of duration \(T_{\mathrm{seg}}\), normalized time is $\tau=\frac{t}{T_{\mathrm{seg}}}\in[0,1]$.

Each segment between consecutive waypoints is then time-scaled by a seventh-order (septic) polynomial profile \(s(\tau):[0,1]\rightarrow[0,1]\) with \(s(0)=0\), \(s(1)=1\), and zero first, second, and third derivatives at both endpoints:
\begin{equation}
\begin{aligned}
\small
\label{eq:septic_profile}
s(\tau)
=
35\tau^4
-
84\tau^5
+
70\tau^6
-
20\tau^7,
\qquad
\tau \in [0,1].
\end{aligned}
\end{equation}

\textcolor{black}{The exact absolute derivative maxima of the normalized profile are}
\begin{equation}
\begin{aligned}
\label{eq:septic_derivative_extrema}
\kappa_v
&=\max_{\tau\in[0,1]}|s'(\tau)|
=\frac{35}{16}=2.1875,\\
\kappa_a
&=\max_{\tau\in[0,1]}|s''(\tau)|
=\frac{84\sqrt{5}}{25}\approx7.51319,\\
\kappa_j
&=\max_{\tau\in[0,1]}|s'''(\tau)|
=\frac{105}{2}=52.5.
\end{aligned}
\end{equation}

For Task~1, let \(q_a^{\mathrm{IK}}\) and
\(q_b^{\mathrm{IK}}\) denote the endpoint-IK solutions of a segment, and
define
\[
\Delta q_j
=
q_{b,j}^{\mathrm{IK}}-q_{a,j}^{\mathrm{IK}},
\qquad
j\in\mathcal{J}_{\mathrm{free}}^{(1)}.
\]
The Task~1 unscaled duration is
\begin{equation}
\begin{aligned}
\bar T_{\mathrm{seg}}^{(1)}
=
\max\Bigg\{&
\max_{j\in\mathcal{J}_{\mathrm{free}}^{(1)}}
\left[
\kappa_v\frac{|\Delta q_j|}{v_{\max}^{(j)}},
\sqrt{\kappa_a\frac{|\Delta q_j|}{a_{\max}^{(j)}}}
\right],
5T_s
\Bigg\}.
\end{aligned}
\label{eq:task1_segment_time_bound}
\end{equation}
No per-joint jerk-limit term is used in this Task~1 rule.

For Task~2, let \(\Delta p_\ell\),
\(\ell\in\{x,y,z\}\), denote the displacement of each Cartesian
position component, and let \(\Delta\theta\) denote the shortest-arc
orientation displacement. The implemented Task~2 unscaled duration is
\begin{equation}
\begin{aligned}
\bar T_{\mathrm{seg}}^{(2)}
=
\max\Bigg\{&
\max_{\ell\in\{x,y,z\}}
\left[
\kappa_v\frac{|\Delta p_\ell|}{v_{\mathrm{xyz}}},
\sqrt{\kappa_a\frac{|\Delta p_\ell|}{a_{\mathrm{xyz}}}},
\sqrt[3]{\kappa_j\frac{|\Delta p_\ell|}{j_{\mathrm{xyz}}}}
\right],\\
&
\kappa_v\frac{\Delta\theta}{\omega_{\max}},
\sqrt{\kappa_a\frac{\Delta\theta}{\alpha_{\max}}},
\sqrt[3]{\kappa_j\frac{\Delta\theta}{j_{\theta,\max}}},
5T_s
\Bigg\}.
\end{aligned}
\label{eq:task2_segment_time_bound}
\end{equation}
The Task~2 Cartesian timing limits are
\[
\begin{aligned}
v_{\mathrm{xyz}}&=0.12~\mathrm{m/s},
&
a_{\mathrm{xyz}}&=0.35~\mathrm{m/s^2},
&
j_{\mathrm{xyz}}&=0.01~\mathrm{m/s^3},\\
\omega_{\max}&=8^\circ/\mathrm{s},
&
\alpha_{\max}&=20^\circ/\mathrm{s^2},
&
j_{\theta,\max}&=60^\circ/\mathrm{s^3}.
\end{aligned}
\]

For \(r\in\{1,2\}\), the implemented duration is scaled and rounded
upward to the controller grid according to
\begin{equation}
T_{\mathrm{seg}}^{(r)}
=
T_s
\left\lceil
\frac{\rho_{\mathrm{pose}}^{(r)}
\bar T_{\mathrm{seg}}^{(r)}}{T_s}
\right\rceil,
\qquad
\rho_{\mathrm{pose}}^{(1)}=1.7,
\quad
\rho_{\mathrm{pose}}^{(2)}=1.9.
\label{eq:scaled_segment_time}
\end{equation}
The Task~2 jerk quantities are Cartesian reference-timing limits, not
per-joint jerk capabilities. These duration calculations specify the
nominal pose reference and do not certify the nonlinear IK solution,
the active-set transitions of the CBF-QP, or the filtered plant
trajectory.

\textcolor{black}{In Task~1, \(q_6\) and \(q_7\) are
time-parameterized by the procedure scheduler. In Task~2, only \(q_7\)
is imposed by the exact scheduler, while \(q_6\) evolves continuously
within the pose-tracking QP.}

\textcolor{black}{Since \(s^{(\ell)}(0)=s^{(\ell)}(1)=0\) for
\(\ell=1,2,3\), consecutive nominal pose-reference segments joined at a common waypoint are \(C^3\), with zero nominal velocity, acceleration, and jerk at the shared waypoint.} This continuity does not transfer automatically to the sampled joint commands because the nonlinear IK map and QP active set can change. Consequently, the sampled joint acceleration may change discontinuously at segment or active-set transitions.

\subsection{Capsule-based self-collision model}

Self-collisions on the SB60 are checked between geometric primitives that approximate the manipulator's bounded regions. Capsules, cylinders with hemispherical end caps, are
a standard abstraction for whole-body manipulator collision
modeling \citep{ragaglia2017cartesian}. In this work, the capsule-pair distance is computed in closed form,
while the distance gradient used in the CBF rows is evaluated by finite
differences with respect to the joint variables. This provides the distance and
local linearization needed for the CBF-QP constraints. We model the SB60 with two related sets of capsule primitives that serve different purposes.

\begin{figure}[h!]
\hspace*{-0.0cm} % Adjust the value as needed
\centering
\scalebox{0.75}{\includegraphics[trim={0cm 0.0cm 0.0cm 0cm},clip,width=\columnwidth]{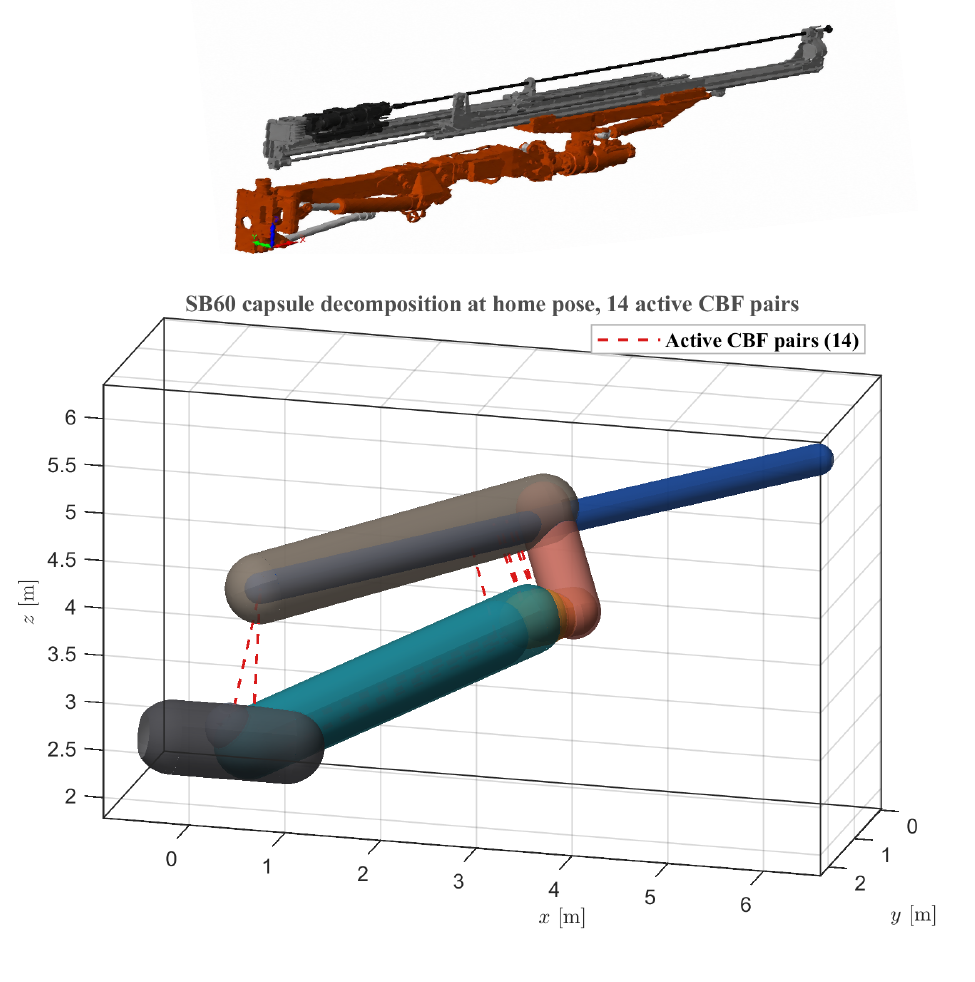}}
\caption{SB60 geometry and the corresponding nine-primitive
whole-body capsule approximation at the home pose. Red dashed segments
indicate the 14 configured capsule pairs.}
\label{fig:capsule_model}
\end{figure}

\textcolor{black}{\textit{1) Whole-body online CBF model: nine primitives and 14 configured pairs.}
The boom is decomposed at the level of major mechanically bounded regions into 9 capsules, listed in Table~\ref{tab:whole_body_capsules}: the carrier base plate, swing cluster, boom-lift assembly, zoom assembly, feed-tilt housing, feed-swing housing, roll/bolt region, cradle, and the long drilling rod. Each capsule axis is anchored to the relevant joint-frame origins or
to the fixed carrier and rig points listed in
Table~\ref{tab:whole_body_capsules}. For example, the boom-lift capsule runs from the swing-cluster end to the boom pivot, and the zoom-assembly capsule runs from the swing-cluster end to the rod cradle, with length increasing as \(q_3\) extends. The configured final capsule radii range from \(0.16~\mathrm{m}\)
for the long rod to \(0.38~\mathrm{m}\) for the carrier base plate.
Of the 36 candidate pair distances among the 9 primitives, eight directly connected neighboring pairs are excluded because their capsules overlap by construction. Fourteen additional pairs are excluded because their guided or intentionally overlapping geometry does not provide a meaningful self-collision constraint in this abstraction; this category includes the nominal overlap between the swing-cluster and cradle capsules. Thus, 22 pairs are excluded in total and the remaining 14 pairs form the online CBF pair set, denoted by \(\mathcal{P}_{\mathrm{CBF}}\). These 14 pair distances are enforced by the CBF-QP and recomputed post-run for controller-set verification. The model is visualized in Fig.~\ref{fig:capsule_model}, and the
14 configured whole-body capsule pairs are listed in
Table~\ref{tab:cbf_pairs}.}

\begin{table}[t]
\centering
\caption{Configured whole-body capsule pairs used by the online CBF-QP.}
\label{tab:cbf_pairs}
\small
\begin{tabular}{cll}
\toprule
Pair index & Primitive \(m\) & Primitive \(n\) \\
\midrule
1  & 1 (\texttt{base\_plate})       & 5 (\texttt{feed\_tilt}) \\
2  & 1 (\texttt{base\_plate})       & 6 (\texttt{feed\_swing}) \\
3  & 1 (\texttt{base\_plate})       & 7 (\texttt{roll\_bolt\_region}) \\
4  & 1 (\texttt{base\_plate})       & 9 (\texttt{long\_rig\_rod}) \\
5  & 2 (\texttt{swing\_cluster})    & 5 (\texttt{feed\_tilt}) \\
6  & 2 (\texttt{swing\_cluster})    & 6 (\texttt{feed\_swing}) \\
7  & 2 (\texttt{swing\_cluster})    & 7 (\texttt{roll\_bolt\_region}) \\
8  & 2 (\texttt{swing\_cluster})    & 9 (\texttt{long\_rig\_rod}) \\
9  & 3 (\texttt{boom\_lift})        & 9 (\texttt{long\_rig\_rod}) \\
10 & 4 (\texttt{zoom\_assembly})    & 9 (\texttt{long\_rig\_rod}) \\
11 & 5 (\texttt{feed\_tilt})        & 8 (\texttt{cradle}) \\
12 & 5 (\texttt{feed\_tilt})        & 9 (\texttt{long\_rig\_rod}) \\
13 & 6 (\texttt{feed\_swing})       & 8 (\texttt{cradle}) \\
14 & 6 (\texttt{feed\_swing})       & 9 (\texttt{long\_rig\_rod}) \\
\bottomrule
\end{tabular}
\end{table}

\begin{table*}[t]
\centering
\caption{Nine-primitive whole-body capsule model used for online CBF
enforcement and post-run controller-set verification. The reported
radii are the effective Task~2 values after configured primitive
inflation.}
\label{tab:whole_body_capsules}
\scriptsize
\begin{tabular}{cllc}
\toprule
\# & Primitive name & Anchor (axis from \(\rightarrow\) to) & Effective radius (m) \\
\midrule
1 & \texttt{base\_plate}        & carrier mount horizontal span                         & 0.38 \\
2 & \texttt{swing\_cluster}     & mount \(\rightarrow\) swing pivot                     & 0.25 \\
3 & \texttt{boom\_lift}         & swing \(\rightarrow\) nominal boom origin             & 0.34 \\
4 & \texttt{zoom\_assembly}     & swing \(\rightarrow\) zoom tip, length grows with \(q_3\) & 0.36 \\
5 & \texttt{feed\_tilt}         & zoom tip \(\rightarrow\) feed-tilt origin             & 0.29 \\
6 & \texttt{feed\_swing}        & feed-tilt \(\rightarrow\) feed-swing origin           & 0.25 \\
7 & \texttt{roll\_bolt\_region} & feed-swing \(\rightarrow\) roll/bolt frame            & 0.28 \\
8 & \texttt{cradle}             & roll/bolt \(\rightarrow\) rig base                    & 0.36 \\
9 & \texttt{long\_rig\_rod} &
rig base\((q_8)\) \(\rightarrow\) rig tip\((q_8)\),
fixed length \(6.7404~\mathrm{m}\) &
0.16 \\
\bottomrule
\end{tabular}
\end{table*}

\textcolor{black}{\textit{2) Offline rod-segment-to-body planning model: 23 selected distances.}
The detour planner uses a reduced model centered on the \(6.74~\mathrm{m}\) drilling rod. The rod is divided into \(n_{\mathrm{seg}}=16\) axial segments and evaluated against seven body-region envelopes. The implementation retains 23 predefined segment-to-region distances for waypoint-risk assessment and candidate scoring.
These 23 local distances are not 23 whole-body capsule pairs and do not imply broader coverage than the 14-pair online model. The planning and CBF models use different primitives and indices, and no containment relation between their modeled clearance sets is assumed. The planning distances are used only to screen offline path segments
and rank detour candidates, whereas the 14 configured CBF pairs include rod-to-body and non-adjacent body-to-body interactions and define \(\mathcal{C}_{\mathrm{CBF}}\). Satisfaction of either model does not imply satisfaction of the other.}

For configured capsule primitives \(m\) and \(n\), let their centerline endpoints be
\(a_m(q),b_m(q),a_n(q),b_n(q)\in\mathbb{R}^3\), with radii \(r_m\) and \(r_n\). Let \(s,t\in[0,1]\) parameterize the two centerline segments. Their signed surface clearance is
\begin{equation}
\label{eq:capsule_pair_clearance}
\begin{aligned}
d_{mn}(q)
=&
\min_{s,t\in[0,1]}
\Big\|
a_m(q)+s\big(b_m(q)-a_m(q)\big)
\\&-
a_n(q)-t\big(b_n(q)-a_n(q)\big)
\Big\|_2
-
(r_m+r_n).
\end{aligned}
\end{equation}

The distance gradient is evaluated using a forward difference:
\[
[\nabla_q d_i^{\mathrm{WB}}(q)]_j
\approx
\frac{
d_i^{\mathrm{WB}}(q+\epsilon_j e_j)
-
d_i^{\mathrm{WB}}(q)
}{
\epsilon_j
},
\]
where
\[
\boldsymbol{\epsilon}_{\mathrm{FD}}
=
[10^{-4},10^{-4},10^{-5},10^{-4},
10^{-4},10^{-4},10^{-4},10^{-5}]^\top.
\]
The increments associated with revolute coordinates are expressed in
radians, whereas those for \(q_3\) and \(q_8\) are expressed in metres.

For each online whole-body pair \(i\in\mathcal{P}_{\mathrm{CBF}}\), let
\(d_i^{\mathrm{WB}}(q)\) denote the signed capsule clearance defined above.
The corresponding online CBF safety margin is
\begin{equation}
\begin{aligned}
\small
\label{eq:capsule_safety_margin}
h_i^{\mathrm{CBF}}(q)
&=
d_i^{\mathrm{WB}}(q)-d_{\mathrm{safe}},
\qquad
i\in\mathcal{P}_{\mathrm{CBF}},
\\
d_{\mathrm{safe}}
&=
0.22~\mathrm{m}.
\end{aligned}
\end{equation}

The value \(d_{\mathrm{safe}}=0.22~\mathrm{m}\) is an engineering
design margin used in this model-based evaluation; it has not been
validated as a physical-machine clearance. The online CBF safe set is

\begin{equation}
\begin{aligned}
\small
\label{eq:capsule_safe_set}
\mathcal{C}_{\mathrm{CBF}}
=
\left\{
q\in\mathbb{R}^{8}
:
h_i^{\mathrm{CBF}}(q)\geq 0,\;
\forall i\in\mathcal{P}_{\mathrm{CBF}}
\right\},
\qquad
\left|\mathcal{P}_{\mathrm{CBF}}\right|=14.
\end{aligned}
\end{equation}
The separate planner distances, denoted by \(d_{\ell}^{\mathrm{plan}}\) for
\(\ell\in\mathcal{P}_{\mathrm{plan}}\) with
\(\left|\mathcal{P}_{\mathrm{plan}}\right|=23\), are used only for offline
waypoint-risk assessment and are not part of
\(\mathcal{C}_{\mathrm{CBF}}\).

\subsection{CBF-constrained QP Control}
\label{sec:cbf_qp_ik}

\textcolor{black}{At each control sample \(k\), the controller uses the measured joint configuration \(q_k\in\mathbb{R}^8\) for forward kinematics, Jacobian evaluation, and capsule-distance calculation. The optimization variable includes all eight joint velocities. Exact equality bounds are imposed only on the joints in \(\mathcal{J}_{\mathrm{exact}}^{(r)}\); therefore, the free velocity variables are those indexed by \(\mathcal{J}_{\mathrm{free}}^{(r)}\). This gives six free joint velocities in Task~1 and seven in Task~2.}

The forward kinematics provide the rig-tip position and orientation,

\begin{equation}
\begin{aligned}
\small
\label{eq:capsule_safasdsadasde_set}
p(q)\in\mathbb{R}^3,\qquad Q(q)\in\mathbb{S}^3,
\end{aligned}
\end{equation}

and the geometric Jacobian maps joint velocity to Cartesian twist,
\begin{equation}
\begin{aligned}
\small
\label{eq:adasdsadasde_set}
\nu =
\begin{bmatrix}
v\\ \omega
\end{bmatrix}
=
J(q)\dot q,\qquad J(q)\in\mathbb{R}^{6\times 8}.
\end{aligned}
\end{equation}

Here, \(\otimes\) denotes scalar-first quaternion multiplication.
The error quaternion is
\[
Q_e=Q_{\mathrm{ref}}\otimes Q(q)^{-1}.
\]
Before applying the logarithm, \(Q_e\) is normalized to unit length and
canonicalized to the nonnegative-scalar hemisphere:
\[
Q_e\leftarrow -Q_e
\qquad\text{if}\qquad
[Q_e]_w<0.
\]
Thus, \( [Q_e]_w\geq0 \), selecting the shortest rotation. For
\(Q_e=[w,\mathbf{v}]\), the implemented logarithm returns the full-angle
rotation vector
\[
\log(Q_e)
=
\begin{cases}
\displaystyle
2\operatorname{atan2}\!\left(\|\mathbf{v}\|_2,w\right)
\frac{\mathbf{v}}{\|\mathbf{v}\|_2},
& \|\mathbf{v}\|_2\geq10^{-12},\\[2ex]
\mathbf{0},
& \|\mathbf{v}\|_2<10^{-12}.
\end{cases}
\]
Accordingly, the returned vector magnitude is the shortest physical
rotation angle, not the half-angle quaternion logarithm. 

All kinematic and distance quantities are evaluated at the measured \(q_k\), closing the controller loop through joint-configuration feedback.

The position and orientation tracking errors are

\begin{equation}
\begin{aligned}
\small
\label{eq:adasdsadasde_set}
e_p =& p_{\mathrm{ref}}-p(q),\\
e_R =&
\log\left(Q_{\mathrm{ref}}\otimes Q(q)^{-1}\right).
\end{aligned}
\end{equation}

where \(\otimes\) denotes quaternion multiplication and \(\log(\cdot)\) maps the unit
quaternion error to its rotation-vector representation. The commanded Cartesian twist is

\begin{equation}
\begin{aligned}
\small
\label{eq:adasdsadasde_set}
\nu_{\mathrm{cmd}} =
\begin{bmatrix}
\dot p_{\mathrm{ref}} + K_p e_p\\
\omega_{\mathrm{ref}} + K_R e_R
\end{bmatrix}.
\end{aligned}
\end{equation}

The QP minimizes soft 6D tracking error, velocity regularization, velocity smoothing,
posture tracking, and CBF slack:
\begin{equation}
\small
\label{eq:adasaadxsasdsadasde_set}
\begin{aligned}
\min_{\dot q,\delta}\quad
&
\frac{1}{2}
\left\|J(q)\dot q-\nu_{\mathrm{cmd}}\right\|_{W_6}^{2}
+
\frac{\lambda}{2}\|\dot q\|_2^2
+
\frac{\beta}{2}\|\dot q-\dot q_{k-1}\|_2^2 \\
&
+
\frac{1}{2}
\left\|\dot q-\dot q_{\mathrm{post}}\right\|_{W_{\mathrm{post}}}^{2}
+
\frac{W_s}{2}\|\delta\|_2^2 ,
\end{aligned}
\end{equation}

subject to

\begin{equation}
\small
\label{eq:adasdsadasde_set}
\begin{aligned}
\dot q_{\min}(q_k,\dot q_{k-1})
\le
\dot q
\le
\dot q_{\max}(q_k,\dot q_{k-1}),
\end{aligned}
\end{equation}

\textcolor{black}{For the task-dependent exact-scheduled joints, the velocity bounds are}
\begin{equation}
\label{eq:scheduled_joint_bounds}
[\dot q_{\min}]_j
=
[\dot q_{\max}]_j
=
\dot q_{j,\mathrm{sch}}^{(r)}(t_k),
\qquad
j\in\mathcal{J}_{\mathrm{exact}}^{(r)}.
\end{equation}
\textcolor{black}{In Task~1 this condition applies to \(q_6\) and \(q_7\), whereas in Task~2 it applies only to \(q_7\). The scheduled joints remain in the Jacobian and capsule-distance gradients, but their velocities cannot be altered by the QP.}

\begin{equation}
\small
\label{eq:online_cbf_constraint}
\begin{aligned}
-\nabla_q d_i^{\mathrm{WB}}(q_k)\dot q-\delta_i
&\le
\alpha h_i^{\mathrm{CBF}}(q_k)
\\
&=
\alpha\left(
d_i^{\mathrm{WB}}(q_k)-d_{\mathrm{safe}}
\right),
\qquad
i\in\mathcal{A}(q_k).
\end{aligned}
\end{equation}

\begin{equation}
\small
\label{eq:adasdsadasde_set}
\begin{aligned}
\delta_i\ge 0,\qquad i\in\mathcal{A}(q_k).
\end{aligned}
\end{equation}

Here, \(\mathcal{A}(q_k)\) is the subset of the 14 online whole-body
pairs that lie within the CBF activation distance:

\begin{equation}
\small
\label{eq:active_online_cbf_set}
\begin{aligned}
\mathcal{A}(q_k)
=
\left\{
i\in\mathcal{P}_{\mathrm{CBF}}
:
d_i^{\mathrm{WB}}(q_k)\leq d_{\mathrm{on}}
\right\},
\qquad
\left|\mathcal{P}_{\mathrm{CBF}}\right|=14.
\end{aligned}
\end{equation}

\textcolor{black}{The posture term provides a soft motion bias for the joints in \(\mathcal{J}_{\mathrm{free}}^{(r)}\). In Task~2, this includes \(q_6\), whose posture reference is soft and does not prevent its use for continuous orientation regulation. Exact-scheduled joints are instead imposed through Eq.~\eqref{eq:scheduled_joint_bounds}.} For Task~2, the posture-reference velocity is
\[
\dot q_{\mathrm{post}}
=
\operatorname{clip}\!\left(
\dot q_{\mathrm{post,ref}}
+
k_{\mathrm{post}}\odot
(q_{\mathrm{post,ref}}-q),
-\dot q_{\mathrm{pull,max}},
\dot q_{\mathrm{pull,max}}
\right),
\]
with
\[
k_{\mathrm{post}}
=
[0.04,0.20,0.30,0.28,0.10,0.02,0.35,0.25],
\]
and
\[
w_{\mathrm{post}}
=
[0.02,0.18,0.28,0.28,0.08,0.01,0.35,0.22].
\]
The posture weight is
\[
W_{\mathrm{post}}
=
g_{\mathrm{post}}\operatorname{diag}(w_{\mathrm{post}}),
\qquad
g_{\mathrm{post}}
=
\min\!\left[
0.30,\,
0.04+0.08g_{\mathrm{slow}}+0.20g_{\mathrm{CBF}}
\right].
\]
The slow-motion thresholds are \(0.003~\mathrm{m/s}\) and
\(0.20^\circ/\mathrm{s}\). The componentwise posture-pull limits are
\[
\dot q_{\mathrm{pull,max}}
=
[0.8^\circ/\mathrm{s},1.2^\circ/\mathrm{s},0.008~\mathrm{m/s},
1.2^\circ/\mathrm{s},0.8^\circ/\mathrm{s},0.6^\circ/\mathrm{s},
1.5^\circ/\mathrm{s},0.008~\mathrm{m/s}]^\top .
\]

The choice \(\alpha h_i^{\mathrm{CBF}}\) is the linear instance of a
general extended class-\(\mathcal{K}\) CBF condition
\(\dot h_i^{\mathrm{CBF}}+
\gamma(h_i^{\mathrm{CBF}})\geq 0\), with
\(\gamma(h_i^{\mathrm{CBF}})
=\alpha h_i^{\mathrm{CBF}}\). The implemented QP can be written in compact form as

\begin{equation}
\small
\label{eq:adasdsadasde_set}
\begin{aligned}
\min_z \frac{1}{2}z^\top H_z z + f_z^\top z,
\end{aligned}
\end{equation}

where \(z=\dot q\) when slack is disabled and
\(z=[\dot q^\top,\delta^\top]^\top\) when slack is enabled. Without slack,

\begin{equation}
\small
\label{eq:adasdsadasde_set}
\begin{aligned}
H = J^\top W_6J + (\lambda+\beta)I + W_{\mathrm{post}},
\end{aligned}
\end{equation}

\begin{equation}
\small
\label{eq:adasdsadasde_set}
\begin{aligned}
f =
-J^\top W_6\nu_{\mathrm{cmd}}
-\beta\dot q_{k-1}
-W_{\mathrm{post}}\dot q_{\mathrm{post}}.
\end{aligned}
\end{equation}

With slack, the CBF inequalities are augmented as

\begin{equation}
\small
\label{esadsaq:adasdsadasde_set}
\begin{aligned}
[A_{\mathrm{CBF}}, -I]
\begin{bmatrix}
\dot q\\
\delta
\end{bmatrix}
\le b_{\mathrm{CBF}}.
\end{aligned}
\end{equation}

Thus, CBF constraints are imposed only for capsule pairs inside the activation distance
\(d_{\mathrm{on}}\). If no capsule pair is inside \(d_{\mathrm{on}}\), the problem reduces to a
bounded least-squares IK QP. \textcolor{black}{When slack is enabled, the numerical optimization vector contains \(8+|\mathcal{A}(q_k)|\) entries. After accounting for the exact scheduled-joint equalities, the number of free decision variables is
\[
|\mathcal{J}_{\mathrm{free}}^{(r)}|+|\mathcal{A}(q_k)|,
\]
namely \(6+|\mathcal{A}(q_k)|\) in Task~1 and \(7+|\mathcal{A}(q_k)|\) in Task~2.}

Because pose tracking is a soft objective, feasibility is determined by the velocity bounds and CBF inequalities. If the componentwise velocity bounds are nonempty, unbounded nonnegative row slacks make the augmented CBF constraints feasible, but positive slack relaxes the corresponding barrier inequality. With \(\lambda>0\), and with \(W_s>0\) when slack is enabled, the
augmented objective is strongly convex and the QP has a unique
minimizer. For the following continuous-time argument, we additionally
assume that the parametric QP solution produces a locally Lipschitz
closed-loop velocity on the evaluated region.

Consider the velocity-level model

\begin{equation}
\small
\label{eq:adasdsadasde_set}
\begin{aligned}
\dot q = u,
\end{aligned}
\end{equation}

with \(u=\dot q^\star\) given by the QP. Assume that every
\(h_i^{\mathrm{CBF}}(q)\), \(i\in\mathcal{P}_{\mathrm{CBF}}\), is continuously
differentiable in the evaluated region, the initial condition satisfies
\(q(0)\in\mathcal{C}_{\mathrm{CBF}}\), the QP is feasible with
\(\delta_i=0\), and the joint velocity applied to the velocity-level model satisfies every relevant
online CBF inequality. Then, for each active online pair
\(i\in\mathcal{A}(q)\),

\begin{equation}
\small
\label{eq:online_cbf_condition}
\begin{aligned}
\dot h_i^{\mathrm{CBF}}(q,\dot q)
+
\alpha h_i^{\mathrm{CBF}}(q)
\geq 0.
\end{aligned}
\end{equation}

On any interval \([t_a,t_b]\) during which
\(i\in\mathcal{A}(q(t))\), the comparison lemma gives

\begin{equation}
\label{eq:online_cbf_comparison_bound}
h_i^{\mathrm{CBF}}(t)
\geq
h_i^{\mathrm{CBF}}(t_a)
e^{-\alpha(t-t_a)},
\qquad
t\in[t_a,t_b].
\end{equation}

Outside the activation set,
\(d_i^{\mathrm{WB}}(q)>d_{\mathrm{on}}>d_{\mathrm{safe}}\), so the barrier margin is strictly positive. Assuming a continuous trajectory, a pair entering the activation set therefore does so with positive margin. Consequently, if the initial configuration lies in
\(\mathcal{C}_{\mathrm{CBF}}\), the barrier gradients used in the inequalities equal the corresponding exact gradients of the modeled capsule-distance functions, and the applied velocity satisfies every relevant zero-slack CBF inequality, the modeled 14-pair safe set is forward invariant for the continuous-time velocity-level model. This result does not apply to the separate 23-distance planning model.

\textcolor{black}{This continuous-time result does not directly extend to the sampled, command-filtered plant loop. No certified intersample bound or sampled-data CBF tightening is implemented \citep{breeden2022sampledCBF}. Positive slack, saturation, acceleration projection, command filtering, finite-difference gradient error, and actuator mismatch can cause the applied velocity to violate an unrelaxed barrier inequality. The result therefore applies only when the applied velocity satisfies every relevant zero-slack constraint. The sampled Simscape histories provide empirical modeled-clearance evidence rather than an intersample or physical-machine safety certificate.}

The CBF-QP and the null-space APF controller evaluated on the same
manipulator in \citep{kivela2018redundant} make different design
trade-offs. In the null-space formulation, collision avoidance is
introduced as a secondary term and therefore depends on the available
redundancy; that method avoids a per-sample QP solve. In the present
method, modeled clearance conditions enter the tracking QP as explicit
inequalities and may modify the tracking motion. Without slack, these
inequalities are hard constraints; with slack, any relaxation is
explicit and penalized. Under the assumptions stated above, the
zero-slack inequalities support a conditional continuous-time
forward-invariance result
\citep{ames2017cbfqp,ames2019cbfTheory}. Neither method provides an
unconditional sampled-data or physical-machine guarantee in the
present comparison. Because target-hardware execution time was not
benchmarked, no deployment-time computational comparison is made.
An APF does not by itself establish forward invariance of a clearance set. The zero-slack CBF inequalities support the conditional continuous-time result stated above, subject to the model, gradient, feasibility, and applied-command assumptions \citep{ames2017cbfqp,ames2019cbfTheory}. Neither formulation provides an unconditional sampled-data or physical-machine guarantee in the present evaluation.

Null-space APF control avoids a per-sample QP solve, whereas the CBF-QP adds this computational cost in exchange for explicit inequality constraints and explicit slack variables. Because execution time on target hardware was not benchmarked, no deployment-time comparison is claimed here.

\begin{itemize}
    \item In the null-space and APF formulation, the safety task is secondary: it acts only through the redundant degrees of freedom and cannot override the primary tracking task. If the desired tracking would violate safety and the redundancy is insufficient to avoid it, the controller can fail into collision or numerical instability. In the CBF-QP formulation, modeled safety is encoded as a primary inequality in the tracking QP. Without slack, the CBF inequalities are hard constraints; with slack enabled, any barrier relaxation is explicit and penalized. The slack mechanism in Eqs.~\eqref{eq:adasaadxsasdsadasde_set} and \eqref{esadsaq:adasdsadasde_set} provides explicit penalized relaxation rather than silent failure.

    \item The null-space and APF formulation relies on a closed-form artificial potential function that produces repulsive forces when the closest distance falls below a threshold. There is no formal forward-invariance guarantee: the controller can in principle approach or cross a safety boundary if the APF gradients are dominated by the primary tracking task.  The CBF-QP formulation provides a conditional continuous-time forward-invariance result under the standard assumptions of the CBF framework \citep{ames2017cbfqp,ames2019cbfTheory}. The implemented sampled-data system is therefore supported by the conditional continuous-time CBF argument for the 14-pair online model and by sampled post-run recomputation of the same modeled margin, rather than by an unconditional sampled-data or physical-machine guarantee.
\end{itemize}

The two formulations are not strictly comparable in their design tradeoffs. Null-space and APF control is computationally lighter, since no QP solve is required, while the CBF-QP provides explicit model-conditional safety constraints at the cost of a per-sample QP solve. For an offline-planned drilling pipeline running at a \(5~\mathrm{ms}\) control rate, the QP cost is acceptable; for a higher-rate manipulator with very tight per-sample compute budgets, null-space and APF control may remain a better choice.

\subsection{Geometric detour-waypoint planner}
\label{sec:geometric_detour_planner}

\textcolor{black}{The four-waypoint chain in Section~\ref{sec:reference_trajectory_generation} can contain segments whose sampled planner clearance falls below
\(d_{\mathrm{safe}}+d_{\mathrm{headroom}}\). 
The geometric planner attempts to improve clearance according to its
separate 23-distance screening model. Because this model differs from
the 14-pair online CBF model, planner acceptance does not guarantee an
improvement in \(h_{\min}^{\mathrm{CBF}}\) or a reduction in CBF
intervention.
Its 23 selected rod-segment-to-body-region distances are used only for offline screening and candidate ranking; they do not define
\(\mathcal{C}_{\mathrm{CBF}}\) or establish forward invariance. For each segment, the planner proceeds as follows.}

\begin{itemize}
\item \textit{Risk evaluation.}
Sample the straight-line segment from \(\mathbf{p}_a\) to \(\mathbf{p}_b\) at \(K_{\mathrm{eval}}=81\) uniformly spaced points. At each sample point, evaluate the 23 selected planner-model distances and compute

\begin{equation}
\label{eq:planner_minimum_distance}
d_{\min}^{\mathrm{plan}}(s_k)
=
\min_{\ell\in\mathcal{P}_{\mathrm{plan}}}
d_{\ell}^{\mathrm{plan}}(s_k),
\qquad
\left|\mathcal{P}_{\mathrm{plan}}\right|=23.
\end{equation}

The distances are obtained either by solving IK at the sample point or, more rapidly, by the reduced geometric evaluation described below. If

\begin{equation}
\begin{aligned}
\small
\label{eq:detour_accept_straight_segment}
\min_k d_{\min}^{\mathrm{plan}}(s_k)
>
d_{\mathrm{safe}} + d_{\mathrm{headroom}},
\end{aligned}
\end{equation}

the segment is accepted by the offline planner.

\item \textit{Safety interpretation.}
This criterion is an offline screening heuristic. The model-conditional
CBF statement applies only to \(\mathcal C_{\mathrm{CBF}}\) under the
assumptions in Section~\ref{sec:cbf_qp_ik}.

\item \textit{Candidate generation.}
If risk is detected, the Task~1 implementation evaluates two
candidate waypoints displaced from a base point along the positive and
negative clearance-normal directions. The retained diagnostic stores
one score for each sign. No 64-candidate random-hemisphere set is
present in the archived run.

    \item \textit{Geometric scoring.}
    For each candidate \(\mathbf{p}_{\mathrm{cand}}\), evaluate the resulting two-sub-segment path
    \begin{equation}
    \begin{aligned}
    \small
    \label{eq:detour_candidate_path}
    \mathbf{p}_a
    \rightarrow
    \mathbf{p}_{\mathrm{cand}}
    \rightarrow
    \mathbf{p}_b
    \end{aligned}
    \end{equation}
    at \(K_{\mathrm{eval}}\) samples per sub-segment. Compute \(d_{\min,\mathrm{cand}}^{\mathrm{plan}}\) as the minimum over the 23 planner-model distances across all sub-segment samples. Score the candidate by maximizing \(d_{\min,\mathrm{cand}}^{\mathrm{plan}}\) across candidates.

    \item \textit{Acceptance.}
    Accept the candidate with the highest score if
    \begin{equation}
    \begin{aligned}
    \small
    \label{eq:detour_candidate_acceptance}
    d_{\min,\mathrm{cand}}^{\mathrm{plan}}
    >
    d_{\mathrm{safe}}
    +
    d_{\mathrm{headroom}} .
    \end{aligned}
    \end{equation}
    If neither evaluated candidate satisfies
Eq.~\eqref{eq:detour_candidate_acceptance}, no planner-approved
single-waypoint detour is available. The highest-scoring rejected
candidate is retained only in the diagnostic record and is not executed.
Instead, the implementation inserts a deterministic raised-transit
fallback containing three waypoints at \(20\%\), \(50\%\), and \(80\%\)
of the original segment. Their height is set to
\[
z_{\mathrm{fb}}
=
\max(z_a,z_b)
+
\max\!\left(
\Delta z_{\mathrm{cross}},
\Delta z_{\mathrm{transit}}
\right),
\]
and their orientations are obtained by shortest-path SLERP at the same
progress values. The resulting waypoint chain is then audited again.
The planner is marked as accepted only if every audited subsegment has
finite clearance satisfying
\(d_{\min}^{\mathrm{plan}}\geq d_{\mathrm{trigger}}\); otherwise, it
returns the fallback path with a warning and
\(\texttt{plan.accepted}=\texttt{false}\).

    \item \textit{Recursion.}
    Apply the algorithm to the resulting two sub-segments. The retained Task~1 configuration used a maximum recursion depth of 1.
\end{itemize}

The retained Task~1 planner configuration used
\(K_{\mathrm{eval}}=81\), two deterministic candidate directions, and
a maximum recursion depth of one. Candidate scoring used
\texttt{evalUseIK=false}; consequently, the 23 planner-model distances
were evaluated directly from the reduced geometric model and no
planner-side IK solves were performed during candidate scoring. The
subsequent execution trajectory was still generated by the
CBF-constrained IK method of Section~\ref{sec:cbf_qp_ik}.

For comparison, applying the full-IK scoring alternative to the same
two-candidate configuration would nominally require

\begin{equation}
N_{\mathrm{IK,nom}}
=
2N_{\mathrm{cand}}K_{\mathrm{eval}}
=
2\cdot2\cdot81
=
324
\end{equation}

IK evaluations per initially risky segment, where the leading factor
of two represents the two candidate subsegments. This count excludes
fallback auditing and subsequent trajectory generation. The previously
reported value of \(4096\) evaluations corresponded to an earlier
\(64\)-candidate, \(32\)-sample configuration that was not used in the
reported Task~1 run.

\begin{figure}[h!]
\hspace*{-0.0cm} % Adjust the value as needed
\centering
\scalebox{1}{\includegraphics[trim={0cm 0.0cm 0.0cm 0cm},clip,width=\columnwidth]{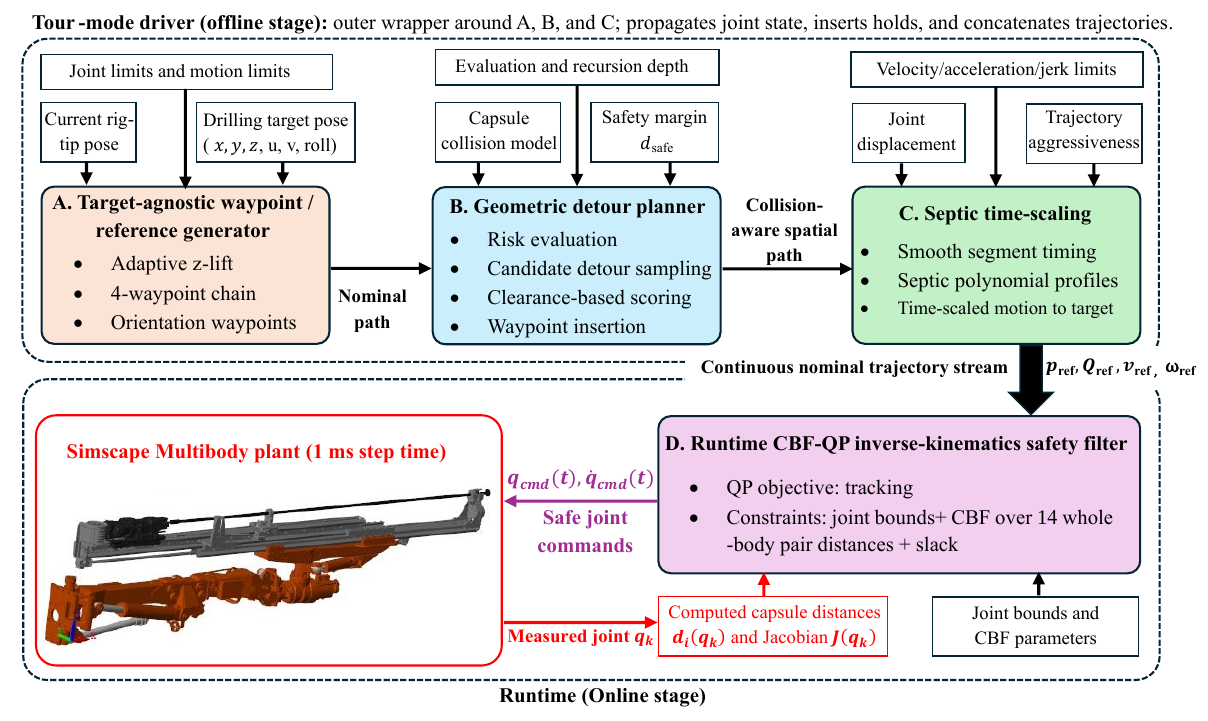}}
\caption{\textbf{Offline planning and closed-loop runtime CBF-QP validation
pipeline.} The offline stage converts each drilling-pose specification into a
continuous nominal reference stream using adaptive waypoint generation,
geometric detour planning, septic time scaling, and tour-mode chaining. The
detour planner evaluates 23 selected rod-segment-to-body-region distances for
offline risk screening and candidate scoring. During runtime, the CBF-QP IK
safety filter receives the reference stream together with the measured joint
state \(q_k\). It recomputes the 14 configured whole-body distances and activates constraints for the subset within \(d_{\mathrm{on}}\)
\(d_i^{\mathrm{WB}}(q_k)\), their gradients, and the geometric Jacobian \(J(q_k)\),
then solves a QP with tracking, joint-bound, CBF, and slack terms. The plant
receives the resulting joint command and returns measured joint states for the
next CBF-QP update. The 23-distance planner audit structures the nominal
path, whereas the 14-pair whole-body CBF model defines the online constraint
set.}
\label{ADADGSFmjhjhjhHG}
\end{figure}

\begin{table*}[t]
\centering
\caption{Task 1 - Drilling-pose set used in the validation experiments.}
\label{tab:drilling_pose_set}
\small
\resizebox{\textwidth}{!}{%
\begin{tabular}{llccccp{4.0cm}}
\toprule
ID & Position \([x,y,z]\) (m) & Drill yaw \(u\) (\(^\circ\)) & Drill pitch \(v\) (\(^\circ\)) & \textcolor{black}{Feed-roll setting \(q_6\)} (\(^\circ\)) & \textcolor{black}{Bolting-tilt setting \(q_7\)} (\(^\circ\)) & Description \\
\midrule
T1 & \([3.25,\;5.50,\;3.80]\) & -90 & 0 & 0 & 90 & Validated baseline \\
T2 & \([4.50,\;3.00,\;3.20]\) & -45 & 0 & 0 & 90 & Mid-front, side angle \\
T3 & \([5.50,\;4.00,\;3.50]\) & -45 & 0 & 0 & 90 & Far-front, high \\
T4 & \([3.00,\;4.50,\;3.50]\) & 0   & 0 & 0 & 90 & Center-front, straight \\
T5 & \([4.50,\;4.50,\;2.50]\) & -30 & 0 & 0 & 90 & Near-front, low \\
\bottomrule
\end{tabular}%
}
\end{table*}

\begin{figure}[t]
\hspace*{-0.0cm} % Adjust the value as needed
\centering
\scalebox{1}{\includegraphics[trim={0cm 0.0cm 0.0cm 0cm},clip,width=\columnwidth]{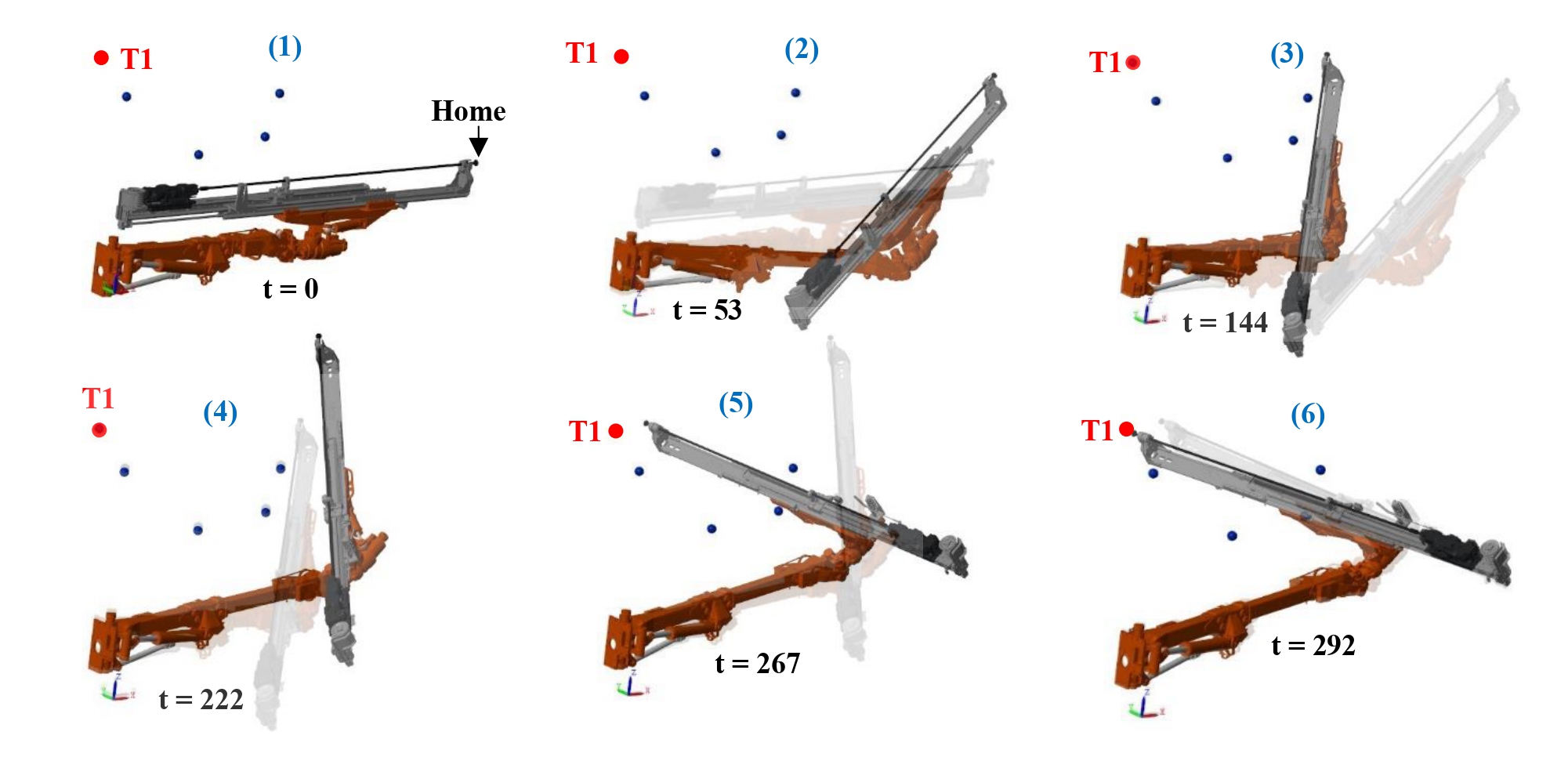}}
\caption{Task~1 Boom motion from Home to T1 using the proposed framework.}
\label{ADADGSFHG}
\end{figure}

\begin{table*}[t]
\centering
\caption{Common and task-specific controller, timing, and
collision-avoidance parameters used in the experiments.}
\label{tab:controller_planner_parameters}
\scriptsize
\begin{tabular}{@{}llcc@{}}
\toprule
Parameter & Symbol & Task~1 & Task~2 \\
\midrule
Controller sample time & \(T_s\) &
\(5~\mathrm{ms}\) & \(5~\mathrm{ms}\) \\
Plant step time & \(T_p\) &
\(1~\mathrm{ms}\) & \(1~\mathrm{ms}\) \\
Position feedback gain & \(K_p\) &
\(0.65\) & \(0.65\) \\
Orientation feedback gain & \(K_R\) &
\(0.45\) & \(0.80\) \\
Position tracking weight & \(W_p\) &
\(1.0\) & \(1.0\) \\
Orientation tracking weight & \(W_R\) &
\(0.10\) & \(1.0\) \\
Velocity regularization weight & \(\lambda\) &
\(10^{-2}\) & \(0.002\) \\
Velocity smoothing weight & \(\beta\) &
\(0.8\) & \(0.10\) \\
Maximum commanded linear speed & \(v_{\mathrm{cmd,max}}\) &
\(0.12~\mathrm{m/s}\) & \(0.12~\mathrm{m/s}\) \\
Maximum commanded angular speed & \(\omega_{\mathrm{cmd,max}}\) &
\(8^\circ/\mathrm{s}\) & \(8^\circ/\mathrm{s}\) \\
CBF safety distance & \(d_{\mathrm{safe}}\) &
\(0.22~\mathrm{m}\) & \(0.22~\mathrm{m}\) \\
CBF activation distance & \(d_{\mathrm{on}}\) &
\(0.45~\mathrm{m}\) & \(0.45~\mathrm{m}\) \\
CBF class-\(\mathcal K\) rate & \(\alpha\) &
\(10\) & \(10\) \\
CBF slack penalty weight & \(W_s\) &
\(10^8\) & \(10^8\) \\
Hold duration & \(T_{\mathrm{hold}}\) &
\(3~\mathrm{s}\) & \(3~\mathrm{s}\) \\
Online CBF pairs & \(N_{\mathrm{CBF}}\) &
\(14\) & \(14\) \\
Offline planner distances & \(N_{\mathrm{plan}}\) &
\(23\) & \(23\) \\
Planner rod segment count & \(n_{\mathrm{seg}}\) &
\(16\) & \(16\) \\
Rod radius & \(r_{\mathrm{rod}}\) &
\(0.16~\mathrm{m}\) & \(0.16~\mathrm{m}\) \\
Rod length & \(L_{\mathrm{rod}}\) &
\(6.7404~\mathrm{m}\) & \(6.7404~\mathrm{m}\) \\
Trajectory timing scale & \(\rho_{\mathrm{pose}}\) &
\(1.7\) & \(1.9\) \\
Cartesian timing speed & \(v_{\mathrm{xyz}}\) &
-- & \(0.12~\mathrm{m/s}\) \\
Cartesian timing acceleration & \(a_{\mathrm{xyz}}\) &
-- & \(0.35~\mathrm{m/s^2}\) \\
Cartesian timing jerk & \(j_{\mathrm{xyz}}\) &
-- & \(0.01~\mathrm{m/s^3}\) \\
Angular timing speed & \(\omega_{\max}\) &
-- & \(8^\circ/\mathrm{s}\) \\
Angular timing acceleration & \(\alpha_{\max}\) &
-- & \(20^\circ/\mathrm{s^2}\) \\
Angular timing jerk & \(j_{\theta,\max}\) &
-- & \(60^\circ/\mathrm{s^3}\) \\
\bottomrule
\end{tabular}
\end{table*}

\begin{figure}[t]
\hspace*{-0.0cm} % Adjust the value as needed
\centering
\scalebox{1}{\includegraphics[trim={0cm 0.0cm 0.0cm 0cm},clip,width=\columnwidth]{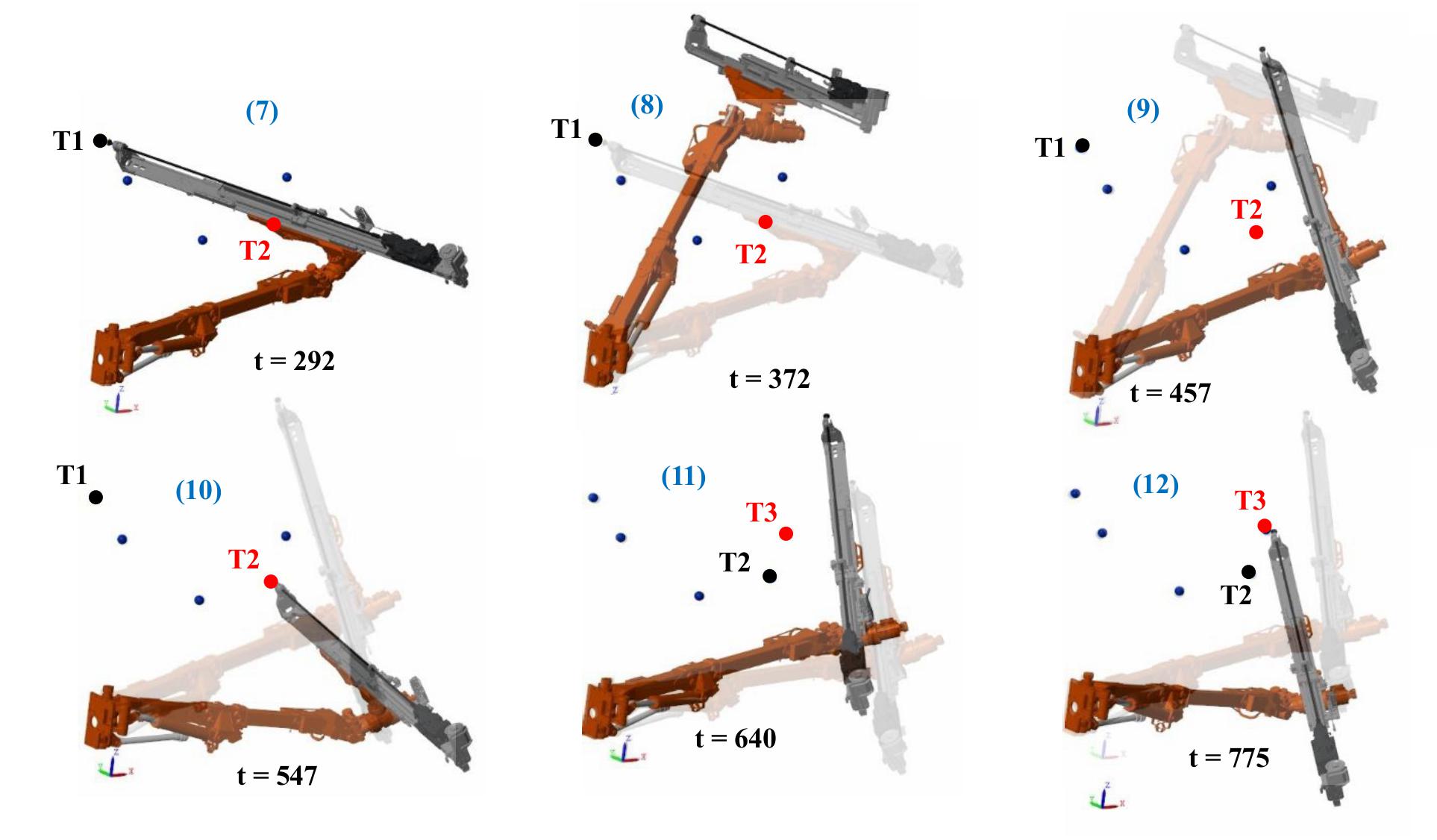}}
\caption{Task~1 Boom motion from T1 to T2 and from T2 to T3 using the proposed framework.}
\label{ADADGSFsadasdaHG}
\end{figure}

\begin{figure}[t]
\hspace*{-0.0cm} % Adjust the value as needed
\centering
\scalebox{1}{\includegraphics[trim={0cm 0.0cm 0.0cm 0cm},clip,width=\columnwidth]{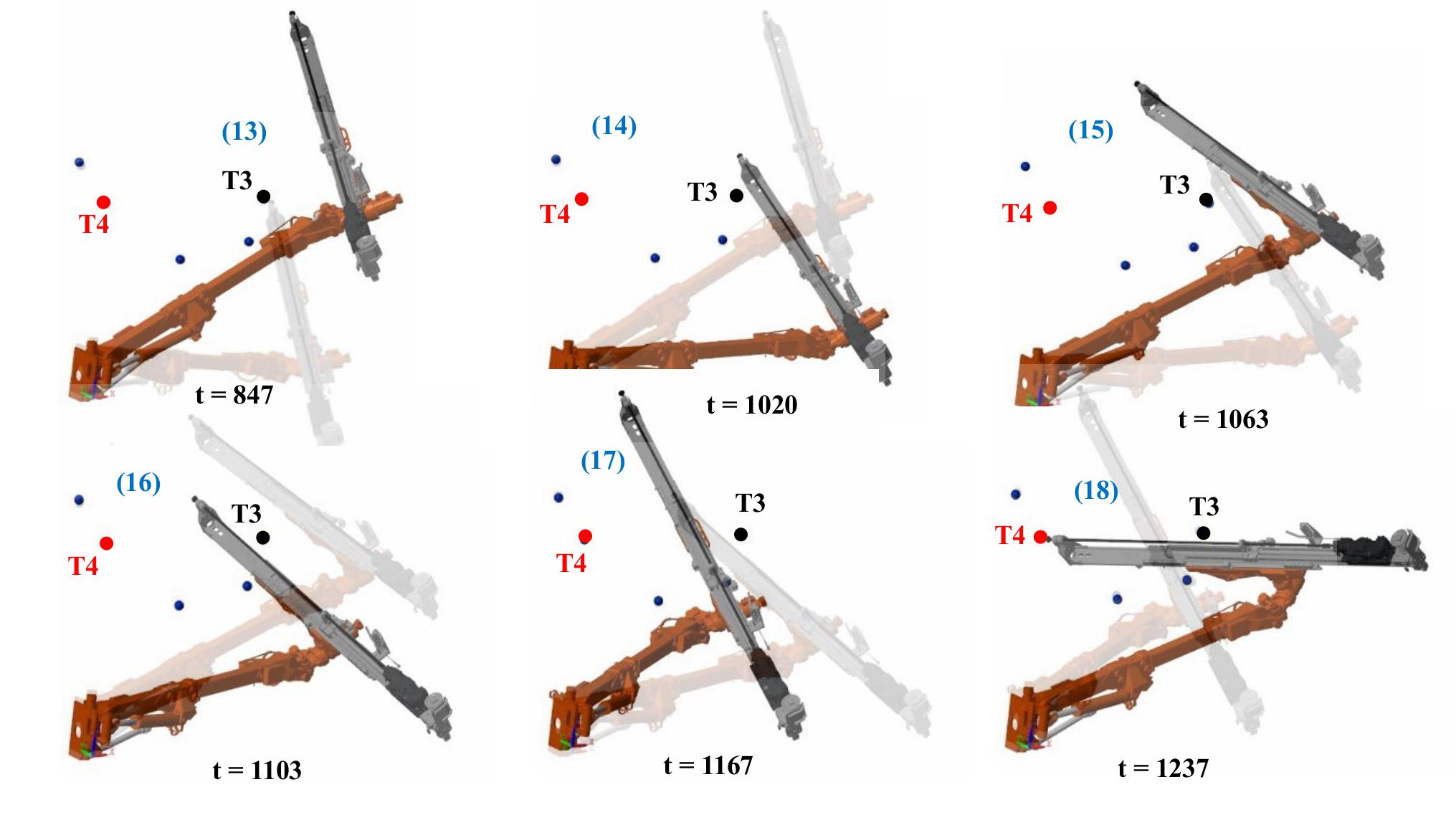}}
\caption{Task~1 Boom motion from T3 to T4 using the proposed framework.}
\label{ADADGhjhjhSFHG}
\end{figure}

\subsection{Tour-mode driver}
\label{sec:tour_mode_driver}

The offline single-pose pipeline produces a trajectory from the rig's current configuration to one drilling pose. To produce a continuous trajectory through a sequence of \(N\) drilling poses, the tour-mode driver runs the pipeline once per hop with three modifications.

\textit{Joint-state continuity.}
The starting joint configuration of hop \(n+1\) is the final joint configuration of hop \(n\):
\begin{equation}
\begin{aligned}
\small
\label{eq:tour_joint_state_continuity}
q_0^{(n+1)}
=
q_f^{(n)},
\qquad
n=1,\ldots,N-1 .
\end{aligned}
\end{equation}
This is what makes the tour an autonomous sequence rather than \(N\) independent point-to-point maneuvers. The rig does not return to a home pose between drilling poses; it transits directly from one drilling configuration to the next.

\textit{Hold insertion.}
After each hop's IK trajectory ends at the drilling pose, the driver appends a hold of duration \(T_{\mathrm{hold}}=3~\mathrm{s}\), during which the commanded joint configuration remains constant at \(q_f^{(n)}\). This allows the plant to settle from the plant-tracking transient at the hop end before the next hop begins. The \(3~\mathrm{s}\) value was chosen empirically based on the plant's approximately \(32~\mathrm{Hz}\) input-filter bandwidth, Section~\ref{sec:plant_model}; shorter holds risk the next hop starting from a non-settled plant state, while longer holds increase tour cycle time without measurable benefit.

\textit{Trajectory concatenation.}
The per-hop IK trajectories, including their appended holds, are concatenated in time. At hop boundary \(n\rightarrow n+1\), the joint position is continuous because \(q_0^{(n+1)}=q_f^{(n)}\), and the nominal joint velocity is zero at the end of the hold and the start of the next transit. The sampled joint acceleration may nevertheless be discontinuous, as discussed in Section~\ref{sec:reference_trajectory_generation}.

Previous studies validated their null-space and APF safety filter on a single point-to-point maneuver, from start pose \(q_s\) to goal pose \(q_g\), with one self-collision avoidance event. The tour-mode driver here generalizes such single-maneuver demonstrations to multi-pose autonomous sequences, where each hop's IK independently solves the safety-constrained tracking problem and the driver chains the hops into a continuous joint command stream. The architecture scales linearly with \(N\): tour build time, IK runtime, and trajectory length all grow proportionally to the number of drilling poses, with no global optimization across hops.

The whole proposed framework including offline planning and closed-loop online CBF-QP controller is shown in Fig. \ref{ADADGSFmjhjhjhHG}.

\begin{figure}[t]
\hspace*{-0.0cm} % Adjust the value as needed
\centering
\scalebox{1}{\includegraphics[trim={0cm 0.0cm 0.0cm 0cm},clip,width=\columnwidth]{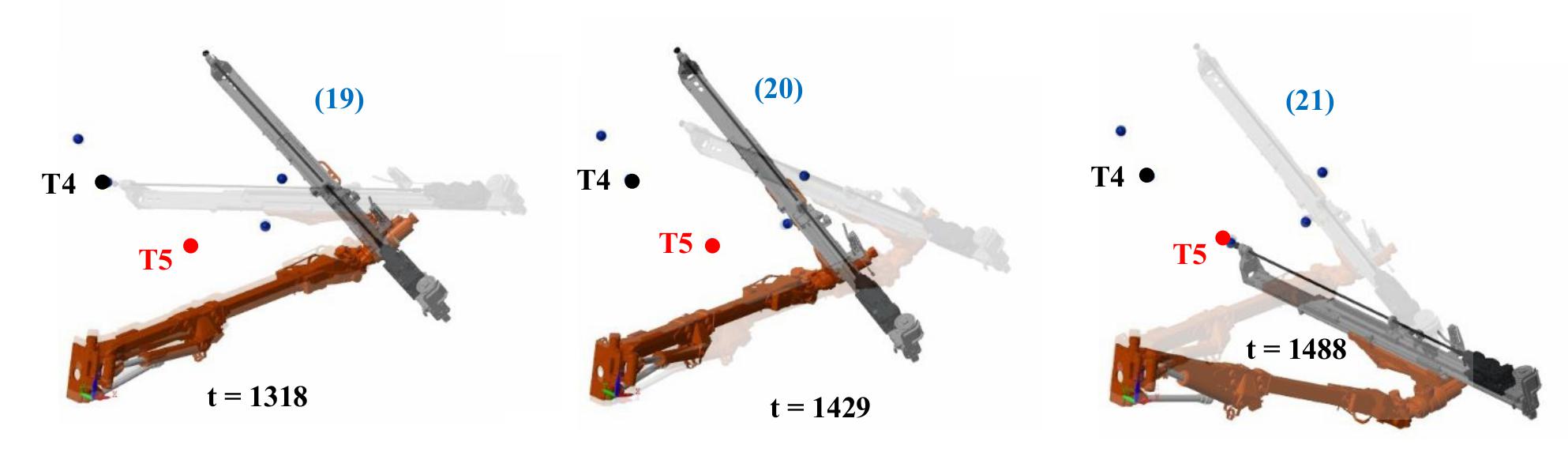}}
\caption{Task~1 Boom motion from T4 to T5 using the proposed framework.}
\label{ADADGhjhjsdsadhSFHG}
\end{figure}

\begin{table}[t]
\centering
\caption{Task 1 - Tracking and safety summary for the full CBF-planner tour.}
\label{tab:full_tour_summary}
\scriptsize
\begin{tabular}{lr}
\toprule
Metric & Value \\
\midrule
Evaluated duration & \(1498.780~\mathrm{s}\) \\
Control samples & \(299{,}757\) \\
Inserted detour waypoints & 13 \\
Planner-triggered transit segments & 11 \\
IK failures & 0 \\
Samples with \(h_{\min}^{\mathrm{CBF}}<-1~\mathrm{mm}\) & 0 \\
Samples with \(h_{\min}^{\mathrm{CBF}}<-5~\mathrm{mm}\) & 0 \\
RMS position error & \(34.08~\mathrm{mm}\) \\
Maximum position error & \(86.19~\mathrm{mm}\) \\
RMS orientation error & \(0.0083^\circ\) \\
Maximum orientation error & \(0.0571^\circ\) \\
Mean hop-final position error & \(4.42~\mathrm{mm}\) \\
Maximum hop-final position error & \(5.99~\mathrm{mm}\) \\
Minimum sampled 14-pair CBF margin \(h_{\min}^{\mathrm{CBF}}\) &
\(-0.002033~\mathrm{mm}\) (\(-2.033~\mu\mathrm{m}\)) \\
Samples with \(h_{\min}^{\mathrm{CBF}}<0\) &
\(8{,}070\) (\(2.692\%\)) \\
\bottomrule
\end{tabular}
\end{table}

\section{\textcolor{black}{Model-based Results}}
\label{sec:experimental_setup}
\textcolor{black}{The framework was evaluated in a
manufacturer-developed SB60 engineering model implemented in Simscape
Multibody. The production controller, hydraulic hardware, physical
sensors, and real boom are outside the evaluation scope.}

The framework was implemented in MATLAB R2024a. The CBF-QP IK solver used \texttt{quadprog} with the active-set algorithm; the detour planner and tour driver were MATLAB scripts wrapped around the solver. The rigid-body Simscape plant applied second-order command filtering at the Simulink-PS Converter blocks with time constant \(T_s=5~\mathrm{ms}\), corresponding to \(f_c\approx(2\pi T_s)^{-1}=31.8~\mathrm{Hz}\).

The Simscape tip-position signal is related to the controller frame by
\begin{equation}
p_{\mathrm{act,sim}}
=
p_{\mathrm{tip,sim}}+\Delta p_{\mathrm{cal}},
\qquad
\Delta p_{\mathrm{cal}}
=
\begin{bmatrix}
-0.0135 & 0.150 & -0.0005
\end{bmatrix}^{\top}\mathrm{m}.
\label{eq:plant_position_calibration}
\end{equation}

The corresponding orientation-frame mapping is
\begin{equation}
\begin{aligned}
Q_{\mathrm{ref,sim}}
&=
Q_{\mathrm{ref,FK}}\otimes Q_{\mathrm{map}},\\
Q_{\mathrm{cmd,sim}}
&=
Q_{\mathrm{FK}}(q_{\mathrm{cmd}})\otimes Q_{\mathrm{map}},\\
Q_{\mathrm{FK,act,sim}}
&=
Q_{\mathrm{FK}}(q_{\mathrm{act}})\otimes Q_{\mathrm{map}},
\end{aligned}
\qquad
Q_{\mathrm{map}}
=
\begin{bmatrix}
0 & \sqrt{0.5} & 0 & \sqrt{0.5}
\end{bmatrix}.
\label{eq:plant_orientation_calibration}
\end{equation}

All quaternions use the \((w,x,y,z)\) convention, and \(Q_{\mathrm{map}}\) is right-multiplied. The calibrated Simscape tip pose is used for the headline post-run tracking metrics; the FK quantities provide consistency diagnostics.

The plant integrates every \(T_p=1~\mathrm{ms}\). Every \(T_s=5~\mathrm{ms}\), the controller reads the simulated joint state, computes a CBF-filtered joint-velocity command, and returns its integrated command to the plant. This closes the loop through simulated joint-state feedback; rig-tip pose is used only for post-run tracking and capsule-margin evaluation.

Table~\ref{tab:controller_planner_parameters} lists the common and task-specific controller parameters. Within each task, these values were fixed across all targets and ablation modes; no target-specific tuning was performed.

\textit{Computational cost.}
The detour planner was executed offline using
\(K_{\mathrm{eval}}=81\), two deterministic candidate directions,
maximum recursion depth one, and geometric candidate scoring
(\texttt{evalUseIK=false}). A controlled wall-time comparison against
full-IK candidate scoring was not retained for this final
configuration; therefore, no planner-runtime speedup is reported. The
CBF-QP was invoked every \(5~\mathrm{ms}\) in simulation.

\subsection{\textcolor{black}{Task 1: Multi-target drilling tour}}

Task~1 evaluates five drilling poses with varying Cartesian position and drill yaw (Table~\ref{tab:drilling_pose_set}). Drill pitch is held at \(0^\circ\), while \(q_6=0^\circ\) and \(q_7=90^\circ\) define the procedure-scheduled feed configuration at the targets. Accordingly, Task~1 evaluates multi-target planning, tracking, and runtime collision filtering, but it is not intended as a full 6-DOF pose-variation test.

With the carrier fixed, the target positions span approximately \(2.5~\mathrm{m}\) in \(x\), \(2.5~\mathrm{m}\) in \(y\), and \(1.3~\mathrm{m}\) in \(z\); several target configurations place joints near their limits. \textcolor{black}{Starting from \(q_{\mathrm{home}}\), the tour visits
\(\mathrm{T1}\rightarrow\mathrm{T2}\rightarrow\mathrm{T3}\rightarrow\mathrm{T4}\rightarrow\mathrm{T5}\),
with \(q_6=0^\circ\) and \(q_7=90^\circ\). Table~\ref{tab:full_tour_summary} summarizes the full-tour results.}

Safety was evaluated using the minimum clearance margin over the 14 configured CBF pairs:
\begin{equation}
\label{eq:cbf_safety_margin_metric}
h_{\min}^{\mathrm{CBF}}(t)
=
\min_{i\in\mathcal{P}_{\mathrm{CBF}}}
\left[
d_i^{\mathrm{WB}}(q(t))-d_{\mathrm{safe}}
\right],
\quad
\left|\mathcal{P}_{\mathrm{CBF}}\right|=14,
\quad
d_{\mathrm{safe}}=0.22~\mathrm{m}.
\end{equation}
A negative \(h_{\min}^{\mathrm{CBF}}\) indicates crossing of at least one modeled pair boundary. Because near-zero negative values may arise from numerical and sampled-data effects, counts below \(-1~\mathrm{mm}\) and \(-5~\mathrm{mm}\) are reported. Unless stated otherwise, all subsequent margins are post-run recomputations of this metric.

Position tracking was evaluated using
\begin{equation}
\label{eq:position_error_metric}
e_{\mathrm{pos}}(t)
=
\left\|
p_{\mathrm{act,sim}}(t)-p_{\mathrm{ref}}(t)
\right\|_2 .
\end{equation}
Trajectory-wide RMS and maximum errors describe transient tracking, whereas mean and maximum hop-final errors describe target-position accuracy.

The shortest-arc quaternion orientation error is
\begin{equation}
e_{\mathrm{ori}}(t)
=
2\cos^{-1}\!\left(
\min\!\left\{
1,
\left|
Q_{\mathrm{ref,sim}}(t)^{\top}
Q_{\mathrm{act,sim}}(t)
\right|
\right\}
\right).
\label{eq:orientation_error_metric}
\end{equation}
The absolute inner product removes quaternion sign ambiguity. The drill-axis pointing error is reported separately because it does not measure rotation about the drill axis.

Figures~\ref{ADADGSFHG}, \ref{ADADGSFsadasdaHG}, \ref{ADADGhjhjhSFHG}, and \ref{ADADGhjhjsdsadhSFHG} show the simulated Home--T1--T2--T3--T4--T5 tour.

\begin{figure}[t]
\centering
\scalebox{0.75}{\includegraphics[trim={0cm 0.0cm 0.0cm 0cm},clip,width=\columnwidth]{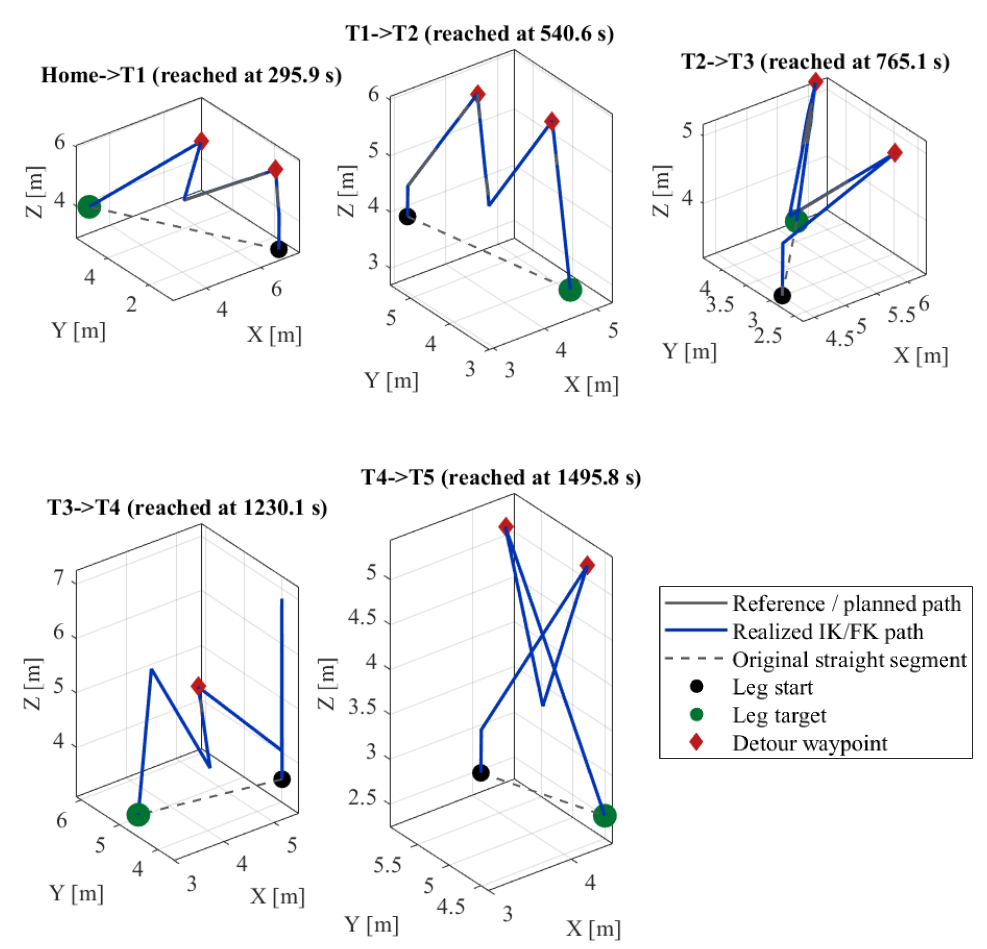}}
\caption{Task 1 - Full-method trajectories by leg. Gray: planned path; blue: realized IK/FK path; dashed gray: original straight segment; black and green markers: leg start and target; red diamonds: inserted detours. Each panel reports its minimum 14-pair CBF margin.}
\label{fig:full_hopwise_trajectory}
\end{figure}

Figure~\ref{fig:full_hopwise_trajectory} compares the planned and realized trajectories by leg.

\begin{table*}[t]
\centering
\caption{Task 1 - Comparative ablation results for the offline detour planner and runtime CBF-QP.}
\label{tab:ablation_comparison}
\scriptsize
\begin{tabular}{lrrrr}
\toprule
Metric & Full &  Planner only & CBF-QP only & Neither \\
\midrule
CBF                  & yes   & no      & yes   & no    \\
Planner              & yes   & yes     & no    & no    \\
\(h_{\min}^{\mathrm{CBF}}\) [mm] & \(\approx 0\) & -332.08 & \(\approx 0\) & -57.26 \\
\(h_{\min}^{\mathrm{CBF}}<-1\) mm & 0 & 28{,}524 & 0 & 4{,}660 \\
\(h_{\min}^{\mathrm{CBF}}<-5\) mm & 0 & 28{,}240 & 0 & 4{,}002 \\
RMS pos. [mm]        & 34.08 & 33.31   & 39.12 & 38.69 \\
Mean hop-final [mm]  & 4.42  & 4.42    & 4.46  & 4.46  \\
\bottomrule
\end{tabular}
\end{table*}

\begin{figure}[t]
\centering
\scalebox{0.75}{\includegraphics[trim={0cm 0.0cm 0.0cm 0cm},clip,width=\columnwidth]{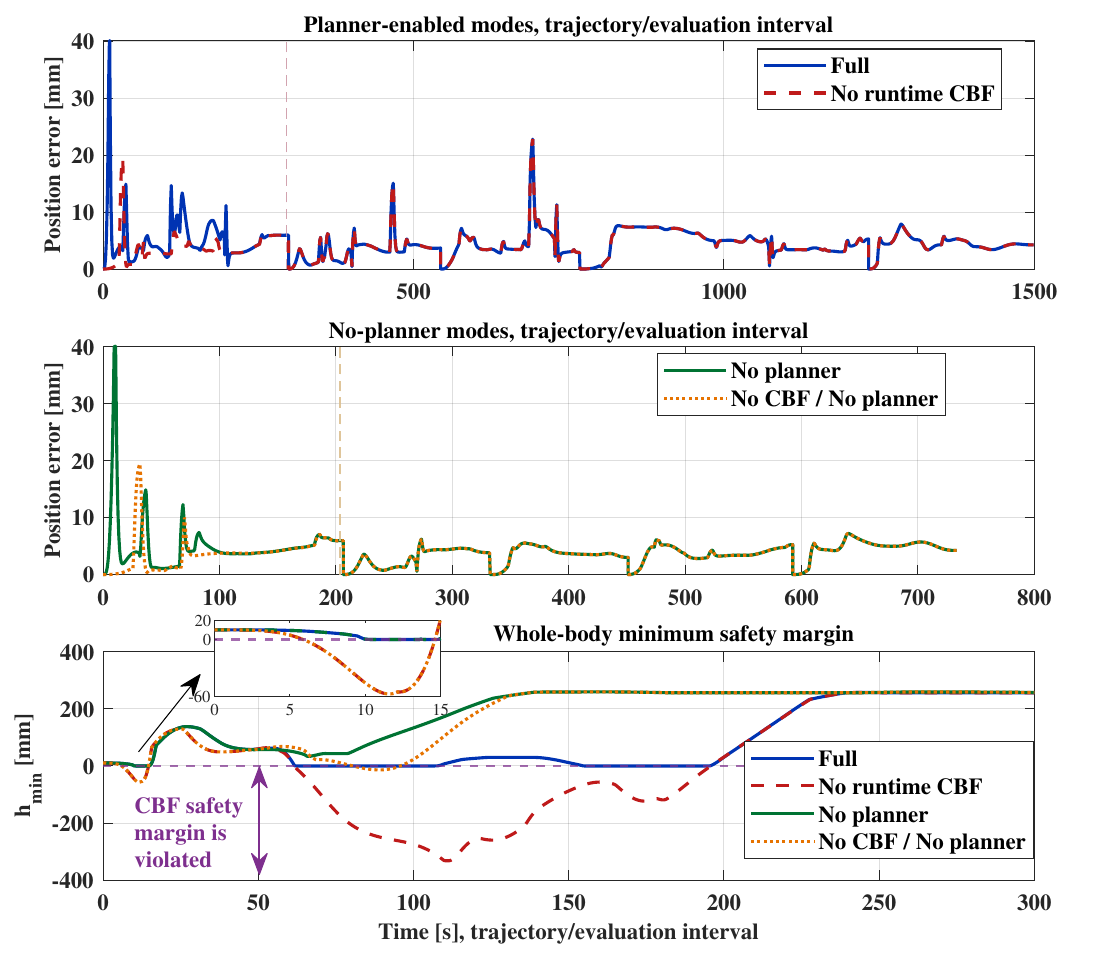}}
\caption{Task 1 - Ablation tracking and safety histories. Top: planner-enabled modes; middle: modes without the planner; bottom: \(h_{\min}^{\mathrm{CBF}}\) for all modes. The dashed line marks \(h_{\min}^{\mathrm{CBF}}=0\); target markers denote Full-mode leg boundaries.}
\label{fig:tracking_safety_grouped}
\end{figure}

Four internal component-ablation modes isolate the offline planner and
runtime CBF-QP: Full, Planner only, CBF-QP only, and Neither. Planner-enabled modes have evaluated durations of approximately \(1498.780~\mathrm{s}\), while modes without the planner have durations of approximately \(733.13~\mathrm{s}\). Figure~\ref{fig:tracking_safety_grouped} groups position histories by trajectory duration and compares all margin histories in the lower panel.

\begin{table}[t]
\centering
\caption{Total comparative orientation-tracking accuracy.}
\label{tab:orientation_summary}
\scriptsize
\begin{tabular}{lrrr}
\toprule
Mode & RMS [deg] & Max [deg] & Final [deg] \\
\midrule
Full proposed method & 0.0083 & 0.0571 & 0.0000 \\
Planner only  & 0.0084 & 0.0571 & 0.0000 \\
CBF-QP only & 0.0098 & 0.0625 & 0.0000 \\
Neither & 0.0099 & 0.0625 & 0.0000 \\
\bottomrule
\end{tabular}
\end{table}

\begin{figure}[t]
\centering
\scalebox{0.75}{\includegraphics[trim={0cm 0.0cm 0.0cm 0cm},clip,width=\columnwidth]{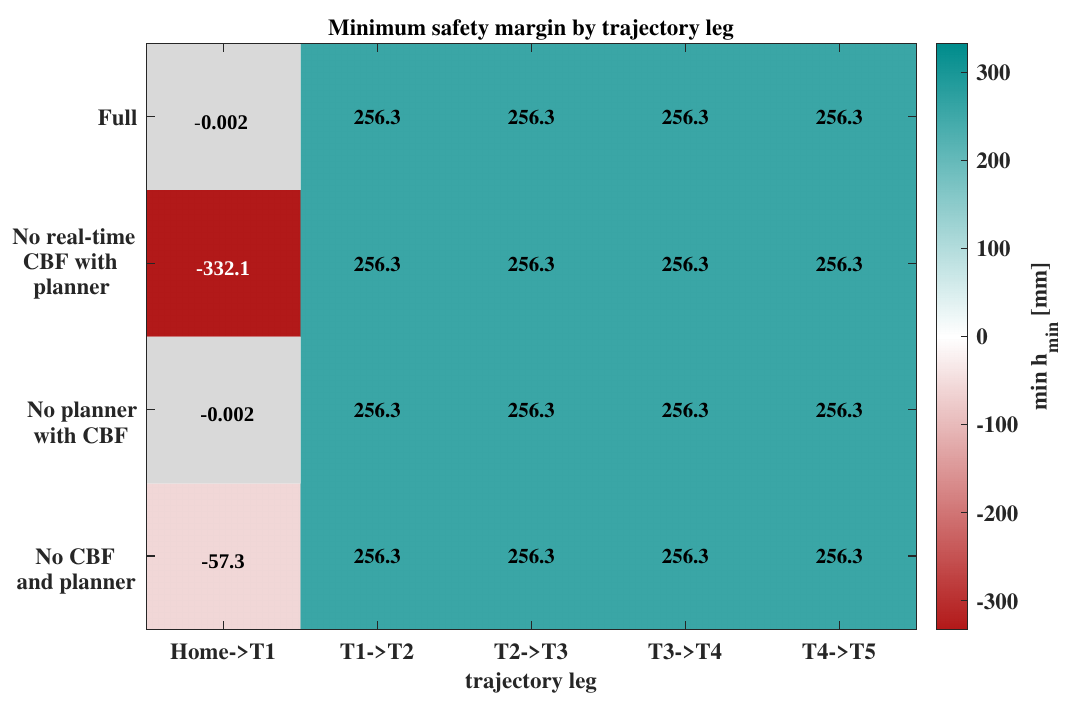}}
\caption{Task 1 - Minimum sampled 14-pair CBF margin by tour leg and ablation mode. Negative values indicate modeled-boundary crossings.}
\label{fig:hopwise_safety_heatmap}
\end{figure}

\begin{figure}[t]
\centering
\scalebox{0.75}{\includegraphics[trim={0cm 0.0cm 0.0cm 0cm},clip,width=\columnwidth]{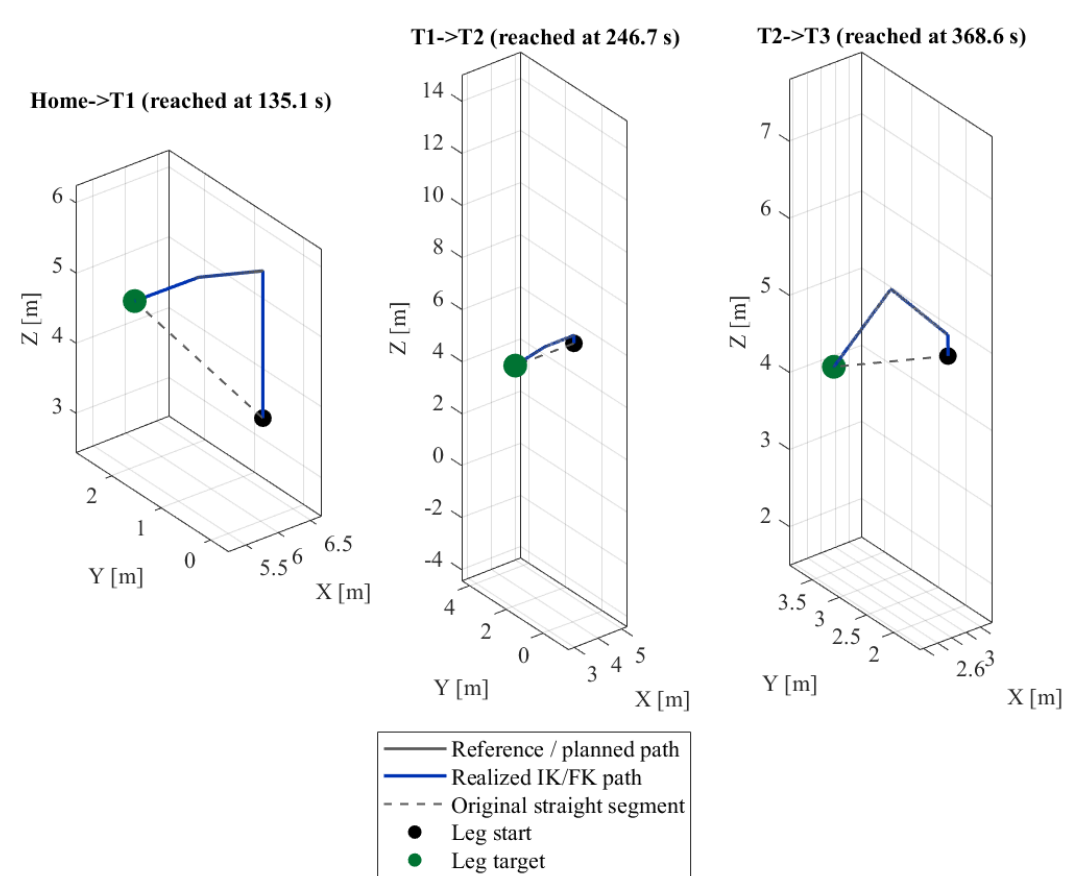}}
\caption{Task~2 full-mode trajectories by leg. Gray: reference path; blue: realized IK/FK path; dashed gray: original straight segment; black and green markers: leg start and target. No geometrically risky segment was detected, and no detour waypoint was inserted.}
\label{fig:task2_hopwise_trajectory}
\end{figure}

In the evaluated tour, both CBF-QP-enabled modes had no samples below the \(-1~\mathrm{mm}\) or \(-5~\mathrm{mm}\) thresholds. Without runtime CBF-QP, Planner only and Neither reached minimum margins of \(-332.08~\mathrm{mm}\) and \(-57.26~\mathrm{mm}\), with \(28{,}240\) and \(4{,}002\) samples below \(-5~\mathrm{mm}\), respectively. Figure~\ref{fig:hopwise_safety_heatmap} localizes these crossings to Home--T1; the inter-target legs retained positive margins. These results apply only to the configured 14-pair capsule model and do not establish clearance for excluded geometry. Across modes, orientation RMS error ranged from \(0.0083^\circ\) to \(0.0099^\circ\), and maximum orientation error ranged from \(0.0571^\circ\) to \(0.0625^\circ\).

Removing the planner while retaining the CBF-QP increased RMS position error from \(34.08~\mathrm{mm}\) to \(39.12~\mathrm{mm}\). The fraction of samples satisfying
\(\min_i d_i^{\mathrm{WB}}\leq d_{\mathrm{on}}\)
increased from \(15.153\%\) with the planner to \(16.871\%\) without it. The lower value with the planner indicates slightly less exposure to the activation region, not lower CBF correction effort; evaluating the latter requires logging binding constraints, slack, and
\(\|\dot q_{\mathrm{safe}}-\dot q_{\mathrm{nom}}\|\).

\subsection{\textcolor{black}{Task 2: Three-Target Full 6‑DOF Drilling-Pose Tracking}}

Task~2 complements the planner- and safety-oriented Task~1 evaluation by varying all six components of the desired end-effector pose. Task~1 varied Cartesian position and drill yaw while retaining fixed pitch and procedure-scheduled feed settings; Task~2 imposes simultaneous variation in \(x\), \(y\), \(z\), yaw, pitch, and roll. The machine starts from the same calibrated home position, \(\mathbf{p}_0=[6.673,\,0.825,\,3.293]^\top~\mathrm{m}\), and visits three drilling poses sequentially in one continuous trajectory.

The target poses are listed in Table \ref{tab:target_poses_6dof}. Positions are expressed in the world frame, and orientations are given as yaw-pitch-roll angles using the ZYX convention.

\begin{table}[t]
\centering
\caption{Task 2 - Cartesian positions and orientations of the target poses.}
\label{tab:target_poses_6dof}
\scriptsize
\begin{tabular}{lrrrrrr}
\toprule
Target & $x$ (m) & $y$ (m) & $z$ (m) & Yaw ($^\circ$) & Pitch ($^\circ$) & Roll ($^\circ$) \\
\midrule
T1 & 5.225 & 1.538 & 5.146 & $-152$ & $-57$ & $25$ \\
T2 & 2.740 & 1.502 & 5.184 & $-121$ & $-46$ & $47$ \\
T3 & 2.981 & 3.922 & 3.851 & $-91$  & $-64$ & $26$ \\
\bottomrule
\end{tabular}
\end{table}

\begin{figure}[t]
\centering
\scalebox{0.75}{\includegraphics[trim={0cm 0.0cm 0.0cm 0cm},clip,width=\columnwidth]{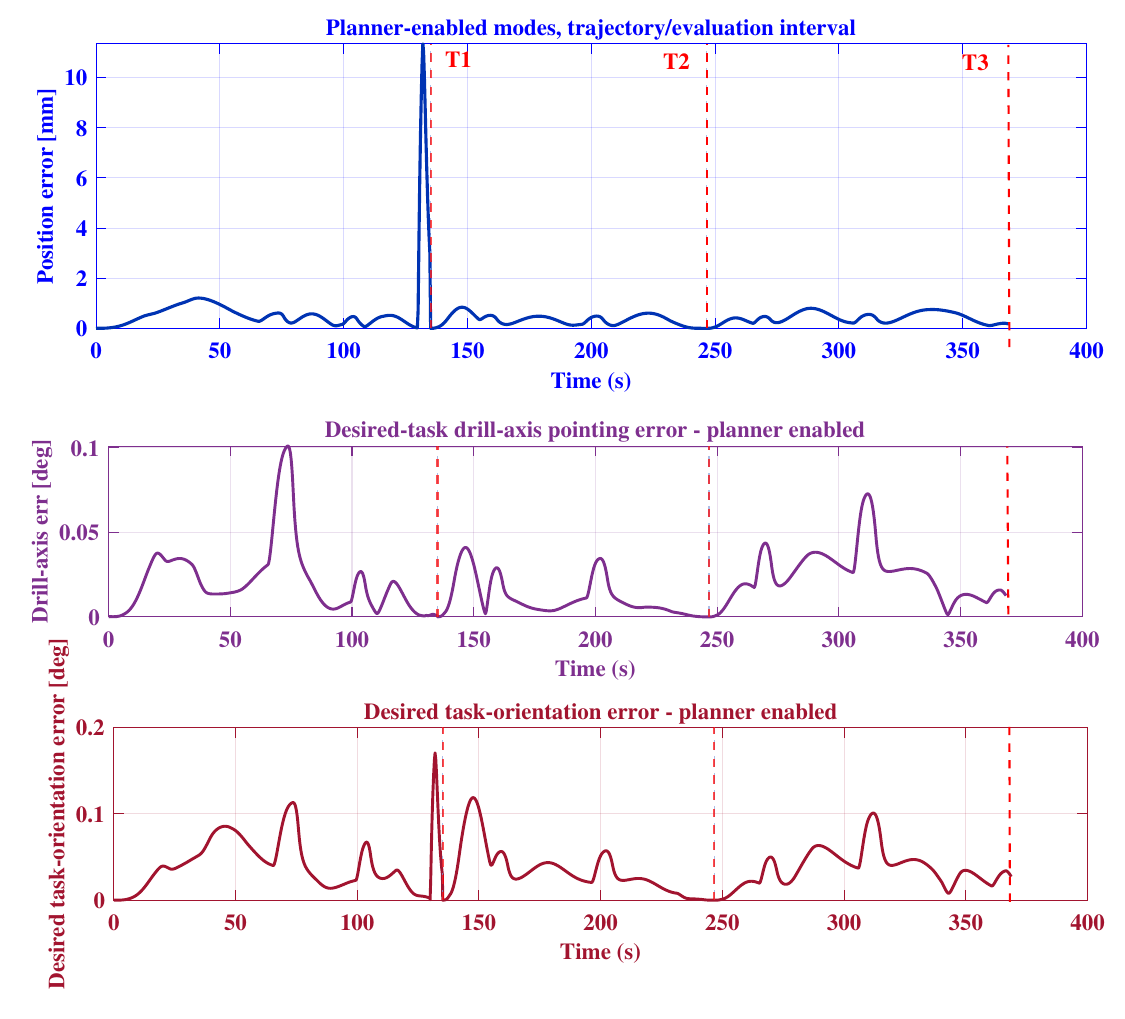}}
\caption{Task~2 full-mode tracking histories. Top: Cartesian position error; middle: drill-axis pointing error; bottom: shortest-arc quaternion orientation error. Vertical dashed lines mark arrival at the three targets.}
\label{fig:task2_tracking_errors}
\end{figure}

Across the target set, the Cartesian coordinates span $2.485\,\mathrm{m}$, $2.420\,\mathrm{m}$, and $1.333\,\mathrm{m}$ in the $x$, $y$, and $z$ directions, respectively. The corresponding orientation spans are $61^\circ$ in yaw, $18^\circ$ in pitch, and $22^\circ$ in roll. Thus, the task explicitly excites all six components of the end-effector pose rather than varying only the drilling position or maintaining a nearly constant tool orientation. Each target was generated from a configuration compatible with the calibrated forward-kinematic model and the configured joint-position limits. The requested poses are therefore kinematically reachable within the evaluated model.

The nominal transit geometry was selected to remain clear of the planner's detour-triggering region. Consequently, none of the three target transitions required an additional geometric detour waypoint. This intentional null planner intervention isolates continuous coupled translational and rotational tracking. The same plant geometry, calibrated home configuration, joint-position ranges, and collision models used in Task~1 were retained; Task~2 used its separately reported full-pose QP weights, timing scale, command bounds, and joint-scheduling configuration.

In Full mode, the \(368.65~\mathrm{s}\), \(73{,}731\)-sample tour achieved position and orientation RMS errors of \(1.005~\mathrm{mm}\) and \(0.0464^\circ\), respectively. The corresponding maximum errors were \(11.342~\mathrm{mm}\) and \(0.1704^\circ\). All samples satisfied the prescribed \(20~\mathrm{mm}\) position and \(2^\circ\) orientation tolerances, yielding a \(100\%\) simultaneous-pose pass rate with zero solver failures.

The terminal position/orientation errors were \(2.50~\mathrm{mm}/0.027^\circ\), less than \(0.001~\mathrm{mm}/0.001^\circ\), and \(0.172~\mathrm{mm}/0.027^\circ\) at the three targets. The minimum sampled surface clearance among the 14 configured online capsule pairs was \(0.251658~\mathrm{m}\), corresponding to a \(31.658~\mathrm{mm}\) modeled margin above \(d_{\mathrm{safe}}=0.22~\mathrm{m}\), with no sampled negative margins.

\begin{figure}[h!]
\hspace*{-0.0cm} % Adjust the value as needed
\centering
\scalebox{1}{\includegraphics[trim={0cm 0.0cm 0.0cm 0cm},clip,width=\columnwidth]{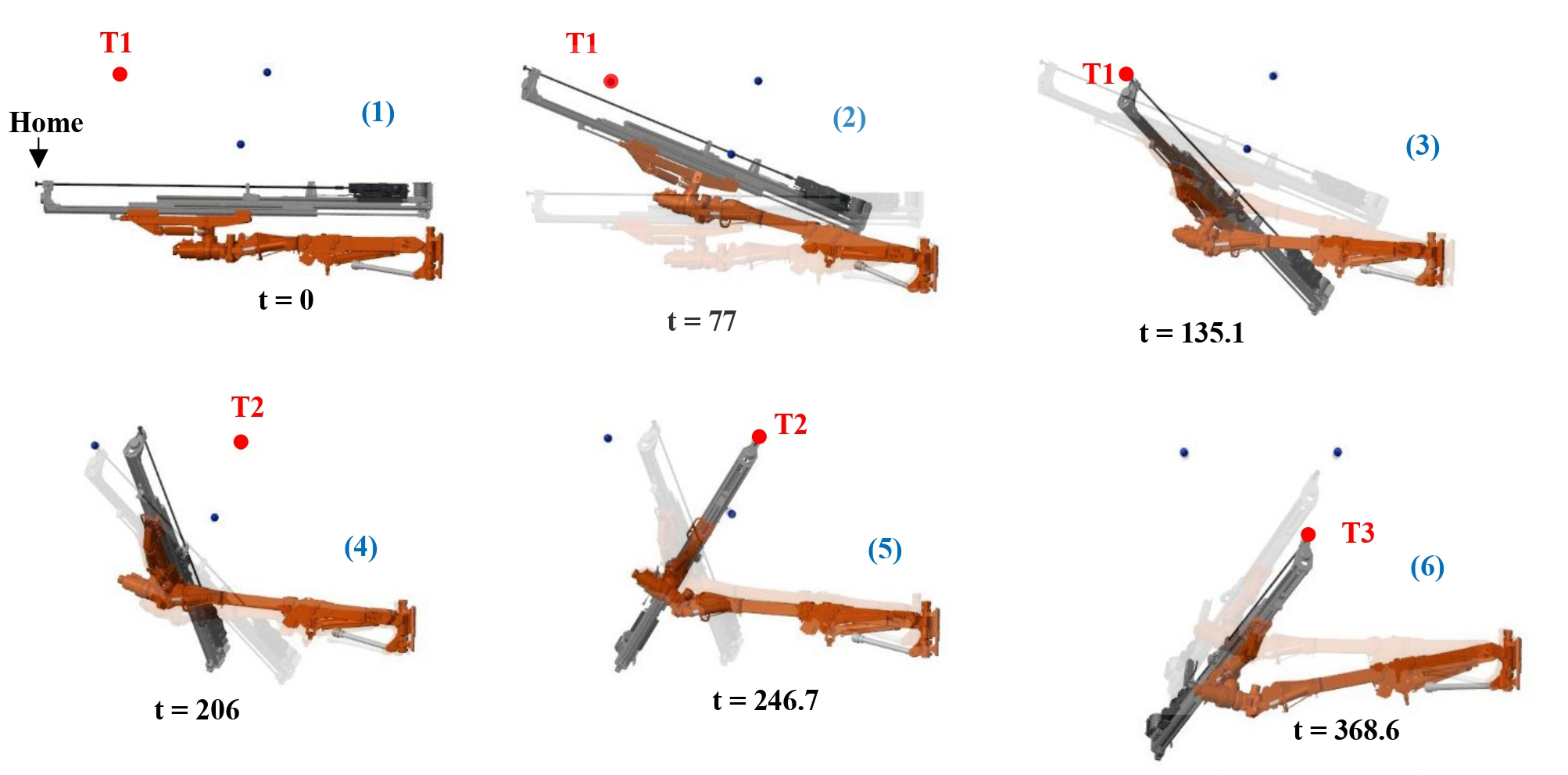}}
\caption{Task~2 Boom motion using the proposed framework.}
\label{fig:task2_boom_motion}
\end{figure}

Fig.~\ref{fig:task2_hopwise_trajectory} shows the realized IK/FK
trajectory closely following the reference path and reaching all three
targets without planner-inserted detours.
Fig.~\ref{fig:task2_tracking_errors} shows consistently low tracking
errors: the maximum position, orientation, and drill-axis errors were
\(11.34~\mathrm{mm}\), below \(0.171^\circ\), and approximately
\(0.10^\circ\), respectively.
Figure~\ref{fig:task2_boom_motion} illustrates the corresponding boom
motion, and Table~\ref{tab:task2_summary} summarizes the full-mode
results.

\begin{table}[t]
\centering
\caption{Task~2 full-mode tracking, planning, and modeled-clearance summary.}
\label{tab:task2_summary}
\small
\begin{tabular}{lr}
\toprule
Metric & Value \\
\midrule
Evaluated duration & \(368.65~\mathrm{s}\) \\
Control samples & \(73{,}731\) \\
Planner-risk segments & 0 \\
Inserted detour waypoints & 0 \\
Solver failures & 0 \\
RMS position error & \(1.005~\mathrm{mm}\) \\
Maximum position error & \(11.342~\mathrm{mm}\) \\
RMS orientation error & \(0.0464^\circ\) \\
Maximum orientation error & \(0.1704^\circ\) \\
Simultaneous-pose pass rate & \(100\%\) \\
Minimum sampled 14-pair clearance & \(0.251658~\mathrm{m}\) \\
Minimum modeled CBF margin & \(31.658~\mathrm{mm}\) \\
Samples with negative CBF margin & 0 \\
\bottomrule
\end{tabular}
\end{table}

The planner-on geometric audit identified no risky transit segment and therefore accepted the nominal trajectory without inserting a detour. Consequently, the Full and CBF-QP-only trajectories were identical. This null planner intervention is expected for the selected geometry
and does not demonstrate independent planner effectiveness. Task~1
instead evaluates the interaction between reduced-model route shaping
and runtime 14-pair filtering; it does not establish the planner as an
independent safety layer.

During Full-mode execution, the runtime CBF-QP remained enabled and the minimum sampled clearance stayed \(31.658~\mathrm{mm}\) above the configured safety threshold. CBF rows entered the activation set during part of the initial motion; however, activation does not necessarily mean that a constraint was binding. Because binding status, slack histories, and \(\|\dot q_{\mathrm{safe}}-\dot q_{\mathrm{nom}}\|\) were not logged, the magnitude of the CBF command modification is not quantified.

\section{Conclusion}

This study integrated pose-parameterized motion generation with measured-state CBF-QP IK for the Sandvik SB60. A fixed task-specific parameter set within each task generated chained
drilling-pose references without pose-specific retuning. The offline planner screened 23 selected rod-segment-to-body-region distances, while the runtime filter imposed 14 configured constraints from a nine-primitive whole-body capsule model.

During a five-target tour evaluated in the manufacturer-developed Simscape Multibody model, the full method inserted 13 detour waypoints and was evaluated over \(299{,}757\) endpoint-inclusive controller samples (\(1498.780~\mathrm{s}\)) with zero IK failures. Mean and maximum hop-final position errors were \(4.42~\mathrm{mm}\) and \(5.99~\mathrm{mm}\), and no sampled CBF margin fell below the \(-1~\mathrm{mm}\) or \(-5~\mathrm{mm}\) thresholds. Without runtime CBF-QP, the planner-only mode reached a minimum margin of \(-332.08~\mathrm{mm}\); with CBF-QP retained, removing the planner increased RMS position error from \(34.08~\mathrm{mm}\) to \(39.12~\mathrm{mm}\). These ablations distinguish online constraint enforcement from offline path shaping.

These findings apply only to sampled evaluations of the configured capsule model and do not establish intersample, excluded-geometry, or physical-machine safety. The continuous-time forward-invariance statement remains conditional on the assumptions in Section~\ref{sec:cbf_qp_ik}. \textcolor{black}{Future work will evaluate the production controller and hydraulic interfaces in hardware-in-the-loop tests before supervised, speed-limited experiments on a physical SB60.}

\section*{{CRediT authorship contribution statement}}
{\textbf{Mehdi Heydari Shahna:} Writing - original draft, Validation, Methodology, Investigation, Formal analysis, Software, Data curation, Conceptualization.}

{\textbf{Tuomo Kivelä:} Writing - review \& editing, Resources, Supervision, Funding acquisition.}

{\textbf{Jouni Mattila:} Writing - review \& editing, Resources, Supervision, Funding acquisition.}

\section*{{Declaration of competing interest}}
{The authors declare that they have no known competing financial interests or personal relationships that could have appeared to influence the work reported in this paper.}

\section*{{Acknowledgement}}
{This work was supported by the Post Docs in Companies (PoDoCo) program in Finland, in collaboration with Tampere University and Sandvik Mining and Rock Technology.}

\bibliographystyle{elsarticle-num} 
\bibliography{manuscript}

\end{document}